\documentclass{article}

\usepackage{xspace,amsmath,amssymb,amsthm,amsfonts,epsfig,syntonly,dsfont,pifont}
\usepackage{natbib}
\usepackage{bm,color,url,textcomp}
\usepackage{algorithmicx,algorithm,algpseudocode}
\usepackage{epstopdf}
\usepackage{empheq,float}
\usepackage{graphicx,subfigure,balance}
\usepackage{multirow}
\usepackage{float}
\usepackage{tcolorbox}

\newtheorem{assumption}{Assumption}

\newtheorem{lemma}{Lemma}
\newtheorem{corollary}{Corollary}
\newtheorem{theorem}{Theorem}
\newtheorem{definition}{Definition}
\theoremstyle{definition}

\DeclareMathOperator*{\argmin}{arg\,min}

\graphicspath{{./figs/}}
\usepackage{wrapfig}

\usepackage{times}
\usepackage{anysize}            % margin package
\marginsize{1.0in}{1.0in}{1in}{1in} % review version with 1 in margins
\usepackage[utf8]{inputenc} % allow utf-8 input
\usepackage[T1]{fontenc}    % use 8-bit T1 fonts
\usepackage{hyperref}       % hyperlinks
\usepackage{url}            % simple URL typesetting
\usepackage{booktabs}       % professional-quality tables
\usepackage{nicefrac}       % compact symbols for 1/2, etc.
\usepackage{microtype}      % microtypography

\usepackage{scrextend}

\title{\huge{Adaptive Step Sizes \\in Variance Reduction via Regularization}}
\author{ 
	Bingcong Li   ~~ Georgios B. Giannakis 	\vspace{0.1cm} \\	 
	 \textit{University of Minnesota - Twin Cities, Minneapolis, MN 55455, USA} \\
	 \texttt{\{lixx5599, georgios\}@umn.edu}
	 }
\vspace{0.3cm} 

\begin{document}

\maketitle
\begin{abstract}
The main goal of this work is equipping convex and nonconvex problems with Barzilai-Borwein (BB) step size. With the adaptivity of BB step sizes granted, they can fail when the objective function is not strongly convex. To overcome this challenge, the key idea here is to bridge (non)convex problems and strongly convex ones via regularization. The proposed regularization schemes are \textit{simple} yet effective. Wedding the BB step size with a variance reduction method, known as SARAH, offers a free lunch compared with vanilla SARAH in convex problems. The convergence of BB step sizes in nonconvex problems is also established and its complexity is no worse than other adaptive step sizes such as AdaGrad. As a byproduct, our regularized SARAH methods for convex functions ensure that the complexity to find $\mathbb{E}[\| \nabla f(\mathbf{x}) \|^2]\leq \epsilon$ is ${\cal O}\big( (n+\frac{1}{\sqrt{\epsilon}})\ln{\frac{1}{\epsilon}}\big)$, improving $\epsilon$ dependence over existing results. Numerical tests further validate the merits of proposed approaches.
\end{abstract}

\section{Introduction}
We study a type of adaptive step size, known as Barzilai-Borwein (BB) step size \citep{barzilai1988}, for solving empirical risk minimization (ERM) problems. The distinct feature of ERM is the finite sum structure of the objective function; that is, 
\begin{align}\label{eq.prob}
	\min_{\mathbf{x} \in \mathbb{R}^d} f(\mathbf{x}) := \frac{1}{n} \sum_{i \in [n]} f_i(\mathbf{x})	
\end{align}
where $\mathbf{x} \in \mathbb{R}^d$ is the parameter vector to be learned from data; the set $[n]:= \{1,2,\ldots, n \}$ collects data indices; and, $f_i$ is the loss function for datum $i$. Throughout, $\mathbf{x}^*$ denotes the optimal solution of \eqref{eq.prob}. It is further assumed that $f(\mathbf{x}^*) > -\infty$. Function $f$ can be convex or nonconvex.

To solve \eqref{eq.prob}, there are basically three types of approaches categorized according to the frequency of computing $\nabla f(\mathbf{x})$. The gradient descent (GD) has a full gradient computed per iteration and updates via
\begin{align*}
	\mathbf{x}_{k+1} = \mathbf{x}_k - \eta\nabla f(\mathbf{x}_k)
\end{align*}
where $k$ is the iteration index and $\eta$ is the step size (or learning rate), see e.g., \citep{nesterov2004}. The stochastic gradient descent (SGD), eliminates the requirement on computing a full gradient by drawing uniformly at random an index $i_k \in [n]$ per iteration $k$, and adopting $\nabla f_{i_k}(\mathbf{x}_k)$ as an unbiased estimate for $\nabla f(\mathbf{x}_k)$; see e.g., \citep{robbins1951,bottou2018,ghadimi2013}. SGD updates are given by
\begin{align*}
	\mathbf{x}_{k+1} = \mathbf{x}_k - \eta_k \nabla f_{i_k}(\mathbf{x}_k)
\end{align*}
where $\eta_k$ is the (diminishing) step size used in iteration $k$. The last type of method is termed variance reduction (VR), which is a family of algorithms that compute $\nabla f(\mathbf{x})$ strategically. The general idea is to judiciously evaluate a \textit{snapshot gradient} $\nabla f(\mathbf{x}_s)$, and use it to form the stochastic gradient estimates in the subsequent iterations. Members of the variance reduction family include  SDCA \citep{shalev2013}, SVRG \citep{johnson2013,reddi2016,allen2016}, SAG \citep{roux2012}, SAGA \citep{defazio2014,reddi2016saga}, MISO \citep{mairal2013}, SARAH \citep{nguyen2017,fang2018}, and their variants \citep{konecny2013,lei2017,kovalev2019}. Most of these algorithms rely on the update 
\begin{align*}
	\mathbf{x}_{k+1} = \mathbf{x}_k - \eta {\cal E}(\nabla f_{i_k}(\mathbf{x}_k), \nabla f(\mathbf{x}_s))
\end{align*} 
where $\eta$ is a constant step size and ${\cal E}(\nabla f_{i_k}(\mathbf{x}_k), \nabla f(\mathbf{x}_s))$ is an algorithm-specific gradient estimate that combines the snapshot gradient $\nabla f(\mathbf{x}_s)$ with the stochastic gradient $\nabla f_{i_k}(\mathbf{x}_k)$.

For the best empirical performance, GD, SGD, and VR entail grid search to tune the step size, which is often painstakingly hard and time consuming. Motivated by the need to alleviate the efforts for tuning, adaptive step sizes for GD and SGD are extensively studied. For example, algorithms such as AdaGrad  \citep{duchi2011,mcmahan2010}, Adam \citep{kingma2014} and their variants \citep{reddi2019,ward2018,li2018convergence,zaheer2018} are popular for training neural networks. While on the other hand, the adaptive step size for VR is not well-documented yet. The first adaptive step size for VR was introduced in \citep{tan2016}, where BB step size for SVRG (BB-SVRG) was studied. BB-SVRG and BB-SARAH were further studied in \citep{liuclass,yang2019,li2019bb}.  However, all these works require a strongly convex $f$. The only work on nonconvex BB step size (that we are aware of) is \citep{ma2018}, modifying the original BB step size for coping with the nonconvexity.

In this work, we answer the following question: \textit{Can BB step sizes be used for convex and nonconvex problems without modification?} The challenge is that strongly convexity is necessary for a proper BB step size, whereas directly employing BB step size in convex problem can lead to an arbitrary large step size; and a negative one in nonconvex problems. Our solution is to bridge (non)convex problems with strongly convex ones through regularization, and then apply BB step size on the strongly convex problems instead. Throughout, we will use SARAH \citep{nguyen2017} as an example, but our methods arguably extend to other VR algorithms such as SVRG.

The proposed regularization approaches are simple and are easy to understand intuitively compared with the more complex regularization schemes in \citep{allen2018}. Our detailed contributions can be summarized as follows. 
\begin{enumerate}
	\item[\textbullet] Convex and nonconvex problems are linked with strongly convex ones through regularization. Our regularization schemes are \textit{simple} in the sense that they only rely on a single quadratic regularizer.
	 
	\item[\textbullet] Multiple regularization techniques are derived for convex and nonconvex problems.	And their corresponding complexities are established. Through regularization, the complexity of SARAH in convex problems is reduced to ${\cal O}\big( (n+\frac{1}{\sqrt{\epsilon}}) \ln \frac{1}{\epsilon} \big)$.
	
	\item[\textbullet] BB step sizes can be then applied on the regularized functions. For convex problems, it is shown that BB step size does not introduce extra complexity compared with vanilla SARAH; while for nonconvex ones, the BB step size is comparable to other adaptive step sizes such as AdaGrad theoretically. The proposed methods are further validated via numerical tests.
\end{enumerate}

\textbf{Notation}. Bold lowercase letters denote column vectors; $\mathbb{E}$ represents expectation; $\| \mathbf{x}\|$ stands for the $\ell_2$-norm of $\mathbf{x}$; and $\langle \mathbf{x}, \mathbf{y} \rangle$ denotes the inner product of vectors $\mathbf{x}$ and $\mathbf{y}$.

%%%%%%%%%%%%%%%%%%%%%%%%%%
\section{Preliminaries}\label{sec.intro}
Basic definitions and assumptions are introduced in this section. Related results for SARAH and BB-SARAH are also reviewed.  

\subsection{Assumptions and Definitions}
The following assumptions will be adopted for convex and nonconvex $f$ with Lipchitz gradient throughout.

\begin{assumption}\label{as.1}
Each $f_i: \mathbb{R}^d \rightarrow \mathbb{R}$ has $L$-Lipchitz gradient; that is, $\|\nabla f_i(\mathbf{x}) - \nabla f_i(\mathbf{y}) \| \leq L \| \mathbf{x}-\mathbf{y} \|, \forall \mathbf{x}, \mathbf{y} \in \mathbb{R}^d$.
\end{assumption} 
\begin{assumption}\label{as.2}
	Each $f_i: \mathbb{R}^d \rightarrow \mathbb{R}$ is convex.
\end{assumption}
Assumption \ref{as.1} requires each loss function to be sufficiently smooth. One can certainly require smoothness of individual loss function and refine Assumption \ref{as.1} as $f_i$ has $L_i$-Lipchitz gradient. Clearly $L =\max_i L_i$. By introducing the importance sampling schemes, such a refined assumption can tighten the $L$ dependence in the bounds. However, since the extension is straightforward and along the lines of those appeared in, e.g., \citep{xiao2014,kulunchakov2019}, we will keep using the simpler Assumption \ref{as.1} for clarity.

As we will link (non)convex problems with strongly convex ones, a formal definition of strong convexity is due.
\begin{definition}
($\mu$-strongly convex.) A continuously differentiable function $f(\mathbf{x})$ is is called $\mu$-strongly convex on $\mathbb{R}^d$ if there exists a constant $\mu > 0$ such that $f(\mathbf{y}) - f(\mathbf{x}) \geq \big\langle  \nabla f(\mathbf{x}), \mathbf{y} - \mathbf{x} \big\rangle	 + \frac{\mu}{2} \| \mathbf{y} - \mathbf{x} \|^2, ~\forall \mathbf{x} \in \mathbb{R}^d, \mathbf{y}\in \mathbb{R}^d$.
\end{definition}
When $f$ is $\mu$-strongly convex with Assumption \ref{as.1} satisfied, the condition number of $f$ is defined as $\kappa = L/\mu$.

Number of iterations for converging is not a fair measure for complexity of an algorithm. Since each iteration of GD requires computing a full gradient, while SGD has much lighter computational burden. In this work the measure for complexity of different algorithms is the incremental first-order oracle (IFO) \citep{agarwal2014}.
\begin{definition}
	An IFO takes $f_i$ and $\mathbf{x} \in \mathbb{R}^d$ as input, and returns the (incremental) gradient $\nabla f_i(\mathbf{x})$.
\end{definition}
For example, computing $\nabla f(\mathbf{x})$ has (IFO) complexity $n$; while $\nabla f_i(\mathbf{x})$ only has complexity $1$. A desirable algorithm obtains a solution with minimal complexity.

\subsection{Recap of SARAH with BB Step Size}\label{sec.recap}
%%%%%%%%%%%%%%%%%%%%%%%%%%%%%%%%%%%%%%%%%%%%%%%%%%%%%%%%%%%
\begin{wrapfigure}{L}{0.5\textwidth}
\begin{minipage}{0.5\textwidth}
\vspace{-0.7cm}
\begin{algorithm}[H]
    \caption{SARAH~/~BB-SARAH}\label{alg.sarah}
    \begin{algorithmic}[1]
    	\State \textbf{Initialize:} $f$, $\tilde{\mathbf{x}}^0 $, $\eta^{(1)}$, $m$, $S$
    	\For {$s=1,2,\dots,S$}
    		\State (\texttt{SARAH}) choose $\eta^{(s)} = \eta^{(1)}$
    		\State (\texttt{BB-SARAH}) choose $\eta^{(s)}$ according to \eqref{eq.bb_stepsize}
			\State $\mathbf{x}_0^s = \tilde{\mathbf{x}}^{s-1}$, and $\mathbf{v}_0^s =  \nabla f (\mathbf{x}_0^s )$
			\State $\mathbf{x}_1^s = \mathbf{x}_0^s - \eta^{(s)} \mathbf{v}_0^s $
			\For {$k=1,2,\dots,m-1$}
				\State uniformly draw $i_k \in [n]$ 
				\State $\mathbf{v}_k^s = \nabla f_{i_k} (\mathbf{x}_k^s ) - \nabla f_{i_k} (\mathbf{x}_{k-1}^s )  + \mathbf{v}_{k-1}^s $
				\State $\mathbf{x}_{k+1}^s = \mathbf{x}_k^s - \eta^{(s)} \mathbf{v}_k^s$
			\EndFor
			\State select $\tilde{\mathbf{x}}^{s}$ randomly from $\{\mathbf{x}_k^s \}_{k=0}^m$ following $\mathbf{p}^s$ 
		\EndFor
		\State \textbf{Return:} $\tilde{\mathbf{x}}^S$
	\end{algorithmic}
\end{algorithm}
\vspace{-0.8cm}
\end{minipage}
\end{wrapfigure}
%%%%%%%%%%%%%%%%%%%%%%%%%%%%%%%%%%%%%%%%%%%%%%%%%%%%%%%%%

The proposed algorithms will rely on the well studied SARAH and BB-SARAH for strongly convex problems. It is thus instructive to recap their convergence properties in both strongly convex and (non)convex cases.

\textbf{SARAH.} The steps of SARAH are listed in Alg. \ref{alg.sarah}. SARAH adopts a biased gradient estimate $\mathbf{v}_k^s$, that is, $\mathbb{E}[\mathbf{v}_k^s|{\cal F}_{k-1}^s]= \nabla f(\mathbf{x}_k^s) - f(\mathbf{x}_{k-1}^s) + \mathbf{v}_{k-1}^s \neq \nabla f(\mathbf{x}_k^s)$, where ${\cal F}_{k-1}^s:= \sigma(\tilde{\mathbf{x}}^{s-1}, i_0, i_1, \ldots, i_{k-1})$. After updating $\mathbf{x}_0^s$ for $m$ times, $\tilde{\mathbf{x}}^s$ is chosen following a pmf vector $\mathbf{p}^s$. There are different options for $\mathbf{p}^s$, and the choice indeed affects the empirical performance of SARAH; see e.g., \citep{nguyen2017,li2019bb}. The complexity of SARAH to ensure $\mathbb{E}[\| \nabla f(\mathbf{x}) \|^2] \leq \epsilon$ is: ${\cal O}\big((n+\kappa) \ln \frac{1}{\epsilon}\big)$ when $f$ is strongly convex; ${\cal O}\big(n+\frac{\sqrt{n}}{\epsilon}\big)$ when $f$ is convex;\footnote{Note that the complexity ${\cal O}\big((n+\frac{1}{\epsilon}) \ln \frac{1}{\epsilon}\big)$ in \citep{nguyen2017}, and ${\cal O}\big(\frac{1}{\epsilon} \ln \frac{1}{\epsilon}\big)$ from \citep{nguyen2018inexact} both require extra assumptions.} and ${\cal O}\big(n+\frac{\sqrt{n}}{\epsilon}\big)$ given a nonconvex $f$. We will henceforth use $\texttt{SARAH}(f,\mathbf{x}_0, \eta, m, S)$ to denote a call of SARAH to minimize $f$, with initial point $\mathbf{x}_0$, and specialized step size $\eta$, inner loop length $m$, and $S$ outer loops.

\textbf{BB-SARAH.} BB step size is helpful for a ``tune-free'' version of SARAH. Suppose that $f$ is $\mu$-strongly convex with condition number $\kappa$. For any $s>1$, BB step size monitors progress of previous outer loops, and chooses the step size of outer loop $s$ accordingly via
\begin{equation}\label{eq.bb_stepsize}
	\eta^{(s)} = \frac{1}{\lambda_\kappa} \frac{\| \tilde{\mathbf{x}}^{s-1} - \tilde{\mathbf{x}}^{s-2} \|^2}{ \big\langle \tilde{\mathbf{x}}^{s-1} - \tilde{\mathbf{x}}^{s-2}, \nabla f(\tilde{\mathbf{x}}^{s-1}) -  \nabla f(\tilde{\mathbf{x}}^{s-2}) \big\rangle}
\end{equation}
where $\lambda_\kappa$ is a $\kappa$-dependent parameter chosen as either $\lambda_\kappa= {\cal O}(\kappa)$ or $\lambda_\kappa= {\cal O}(\kappa^2)$; see \citep{tan2016,liuclass,li2019bb} for details. Note that $\nabla f(\tilde{\mathbf{x}}^{s-1})$ and $\nabla f(\tilde{\mathbf{x}}^{s-2})$ are computed at the outer loops $s$ and $s-1$, respectively; hence, the BB step size only introduces almost negligible memory overhead to store $\tilde{\mathbf{x}}^{s-2}$ and $\nabla f(\tilde{\mathbf{x}}^{s-2})$ in implementation. The price paid for an automatically tuned BB step size is a worse $\kappa$ dependence in the complexity, i.e., ${\cal O}\big( (n+\kappa^2)\ln \frac{1}{\epsilon}\big)$. For convenience, we will use $\texttt{BB-SARAH}(f,\mathbf{x}_0, \eta^{(0)}, m, S)$ to denote a call of SARAH with BB step size to minimize $f$, with initial point $\mathbf{x}_0$, and initialized step size $\eta^{(0)}$, inner loop length $m$ and the number of outer loops $S$. 

Strongly convexity on the objective function plays an important role for a properly defined BB step size. One can verify that the BB step size in \eqref{eq.bb_stepsize} lies in the following interval given a strongly convex $f$ \citep{tan2016,liuclass,li2019bb}
\begin{align*}
	\frac{1}{\lambda_\kappa L} \leq \eta^s \leq \frac{1}{\lambda_\kappa \mu}.
\end{align*}
Such bounds on $\eta^s$ break down when $f$ is (non)convex. Specifically, the denominator of $\eta^{s}$ in \eqref{eq.bb_stepsize} can approach to $0$ or even become a negative number for convex and nonconvex $f$, respectively. Such behavior is observed in practice as well \citep{ma2018} .

Our goal in this paper is to equip convex and nonconvex SARAH with BB step size. To overcome the aforementioned challenge, the general idea is to add a quadratic regularizer to $f$ for obtaining a strongly convex surrogate function $\tilde{f}$, and then apply BB step size on $\tilde{f}$. The convergence of proposed approaches will be established by exploring the connections between $\tilde{f}$ and $f$. Note that the BB step sizes for convex and nonconvex problems were studied in \citep{ma2018}. Our proposed approach enjoys a lower complexity than \citep{ma2018} in convex problems. And combining our regularization techniques with \citep{li2019bb}, it is possible to obtain a ``tune-free'' version of BBstep size for (non)convex problems.

\section{Bridge (Non)Convexity and Strongly Convexity}\label{sec.cvx}
To markedly broaden the scope fo BB step sizes to (non)convex problems, the general idea is to define a surrogate function by adding a quadratic regularizer to $f$. Regularization brings strongly convexity to the problem, which in turn implies that: i) the surrogate function is easier to be optimized compared with $f$; ii) BB step size can be directly applied on the surrogate function. This section develops general regularization schemes for SARAH.

\subsection{Convex Case}
Convex problems satisfying Assumptions \ref{as.1} and \ref{as.2} will be treated in this subsection. Intuitively, the surrogate function obtained through adding a quadratic regularizer scaled by a small number would not deviate from $f$ that much. Hence optimizing it yields a solution approaching to $\mathbf{x}^*$. In addition, strong convexity of the surrogate function will lead to linear convergence rate. It is thus natural to expect improvement on complexity bounds. 

\noindent
%%%%%%%%%%%%%%%%%%%%%%%%%%%%%%%%%%%%%%%%%%%%%%%%%%%%%%%%%%%
\begin{minipage}[t]{0.5\textwidth}
\vspace{-0.5cm}
\begin{algorithm}[H]
    \caption{Regularized SARAH}\label{alg.single_reg}
    \begin{algorithmic}[1]
    	\State \textbf{Initialize:} $\mathbf{x}_0$
    	\State define $\tilde{f}$ as in \eqref{eq.reg_f}
		\State $\tilde{\mathbf{x}} \leftarrow \texttt{SARAH}(\tilde{f}, \tilde{\mathbf{x}}_{0}, \eta, m, K)$ 
		\State \textbf{Return:} $\tilde{\mathbf{x}}$
	\end{algorithmic}
\end{algorithm}
\end{minipage}
%%%%%%%%%%%%%%%%%%%%%%%%%%%%%%%%%%%%%%%%%%%%%%%%%%%%%%%%%
%%%%%%%%%%%%%%%%%%%%%%%%%%%%%%%%%%%%%%%%%%%%%%%%%%%%%%%%%%%
\begin{minipage}[t]{0.5\textwidth}
\vspace{-0.5cm}
\begin{algorithm}[H]
     \caption{Recursively Regularized SARAH (convex)}\label{alg.recursive_sarah}
    \begin{algorithmic}[1]
    	\State \textbf{Initialize:} $\mathbf{x}_0$, $\tilde{\mathbf{x}}_0 \leftarrow \mathbf{x}_0$
    	\For {$s=1,2,\dots,S$}
    		\State (v1) define $\tilde{f}_s$ as in \eqref{eq.rr_reg}
    		\State (v2) define $\tilde{f}_s$ as in \eqref{eq.rr_reg2}
			\State $\tilde{\mathbf{x}}_s \leftarrow \!\texttt{SARAH}(\tilde{f}_s, \tilde{\mathbf{x}}_{s-1}, \eta_s, m_s, K_s)$ 
		\EndFor
		\State \textbf{Return:} $\tilde{\mathbf{x}}^S$
	\end{algorithmic}
\end{algorithm}
\end{minipage}
%%%%%%%%%%%%%%%%%%%%%%%%%%%%%%%%%%%%%%%%%%%%%%%%%%%%%%%%%

\subsubsection{SARAH with A Single Regularizer}
The simplest regularization scheme one can think of is just adding a single quadratic function to $f$. Surprisingly, such simple regularization leads to significant improvement on complexity. Define the surrogate function as
\begin{align}\label{eq.reg_f}
	\tilde{f}(\mathbf{x}) & = f(\mathbf{x}) + \frac{\mu}{2} \| \mathbf{x} - \mathbf{x}_0 \|^2   = \frac{1}{n}\sum_{i\in [n]}f_i(\mathbf{x}) + \frac{\mu}{2} \| \mathbf{x} - \mathbf{x}_0 \|^2	
\end{align}
where $\mathbf{x}_0$ is a given point. It can be verified that $\tilde{f}$ is $L+\mu$ smooth, and $\mu$ strongly convex when Assumptions \ref{as.1} and \ref{as.2} hold. The condition number of $\tilde{f}$ is $\tilde{\kappa}:= (L+\mu)/\mu$. One can also rewrite $\tilde{f}(\mathbf{x}) =  \frac{1}{n}\sum_{i\in [n]} \big[ f_i(\mathbf{x}) + \frac{\mu}{2} \| \mathbf{x} - \mathbf{x}_0 \|^2 \big]:=  \frac{1}{n}\sum_{i\in [n]} \tilde{f}_i(\mathbf{x})$. Clearly, $\tilde{f}_i$ is $\mu$-strongly convex as well. The strongly convexity of $\tilde{f}_i$ can be helpful for choosing $\mathbf{p}^s$ in SARAH \citep{li2019bb}. Leveraging $\tilde{f}$, the regularized SARAH is summarized in Alg. \ref{alg.single_reg}, and its complexity bound is established next.

\begin{theorem}\label{thm.single_reg}
	Suppose that Assumptions \ref{as.1} and \ref{as.2} are satisfied. Choosing $\mu = {\cal O}\big(\frac{\sqrt{\epsilon}}{\|\mathbf{x}_0 - \mathbf{x}^* \|}\big)$, $\eta<\frac{1}{\mu+L}$, $m = {\cal O}\big(\frac{L+\mu}{\mu}\big)$ and $K = {\cal O} \big(\ln \frac{\| \nabla f(\mathbf{x}_0) \|^2}{\epsilon} \big)$, Alg. \ref{alg.single_reg} can obtain $\mathbb{E} \big[ \| \nabla f(\mathbf{x}) \|^2\big] \leq \epsilon$ with complexity
	\begin{align*}
		{\cal O} \bigg( \Big(n+  \frac{L \| \mathbf{x}_0 - \mathbf{x}^* \|}{ \sqrt{\epsilon}} \Big) \ln \frac{\| \nabla f(\mathbf{x}_0) \|^2}{\epsilon} \bigg).
	\end{align*}
\end{theorem}

The complexity bound in Theorem \ref{thm.single_reg} improves at least the $\epsilon$ dependence over all other SARAH variants for convex problems in the literature. So far, the complexity of SARAH for finding $\mathbb{E}\big[ \| \nabla f(\mathbf{x})\|^2 \big] \leq \epsilon$ is ${\cal O}\big(n+\frac{\sqrt{n}}{\epsilon} \big)$ \citep{li2019l2s}.\footnote{Note that the complexity ${\cal O}\big((n+\frac{1}{\epsilon}) \ln \frac{1}{\epsilon}\big)$ in \citep{nguyen2017}, and ${\cal O}\big(\frac{1}{\epsilon} \ln \frac{1}{\epsilon}\big)$ from \citep{nguyen2018inexact} both require extra assumptions. Hence comparing those bounds to Theorem \ref{thm.single_reg} is not fair, even though Theorem \ref{thm.single_reg} still improves over these works.} Comparing now with the complexity of vanilla SARAH, the bound in Theorem \ref{thm.single_reg} depends on the summation of $n+\frac{1}{\sqrt{\epsilon}}$ rather than their product. This means that even a large $n$ only has mild influence on the complexity. Note that our goal here is to find $\mathbb{E}[\| \nabla f(\mathbf{x}) \|^2] \leq \epsilon$, which is easier than $ \mathbb{E}\big[f(\mathbf{x}) - f(\mathbf{x}^*)\big] \leq \epsilon$, since $\frac{1}{2L}\| \nabla f(\mathbf{x}) \|^2 \leq f(\mathbf{x}) - f(\mathbf{x}^*)$ holds when $f$ is convex. Our result is thus not violating on the lower bound on $f(\mathbf{x}) - f(\mathbf{x}^*)$ established in \citep{woodworth2016}.

\subsubsection{SARAH with Recursive Regularizer v1}

\begin{wrapfigure}{r}{0.38\textwidth}
\vspace{-0.7cm}
\begin{tcolorbox}
	\centering
	\hspace{-0.2cm}
	\vspace{-0.3cm}
	\includegraphics[height=3.3cm]{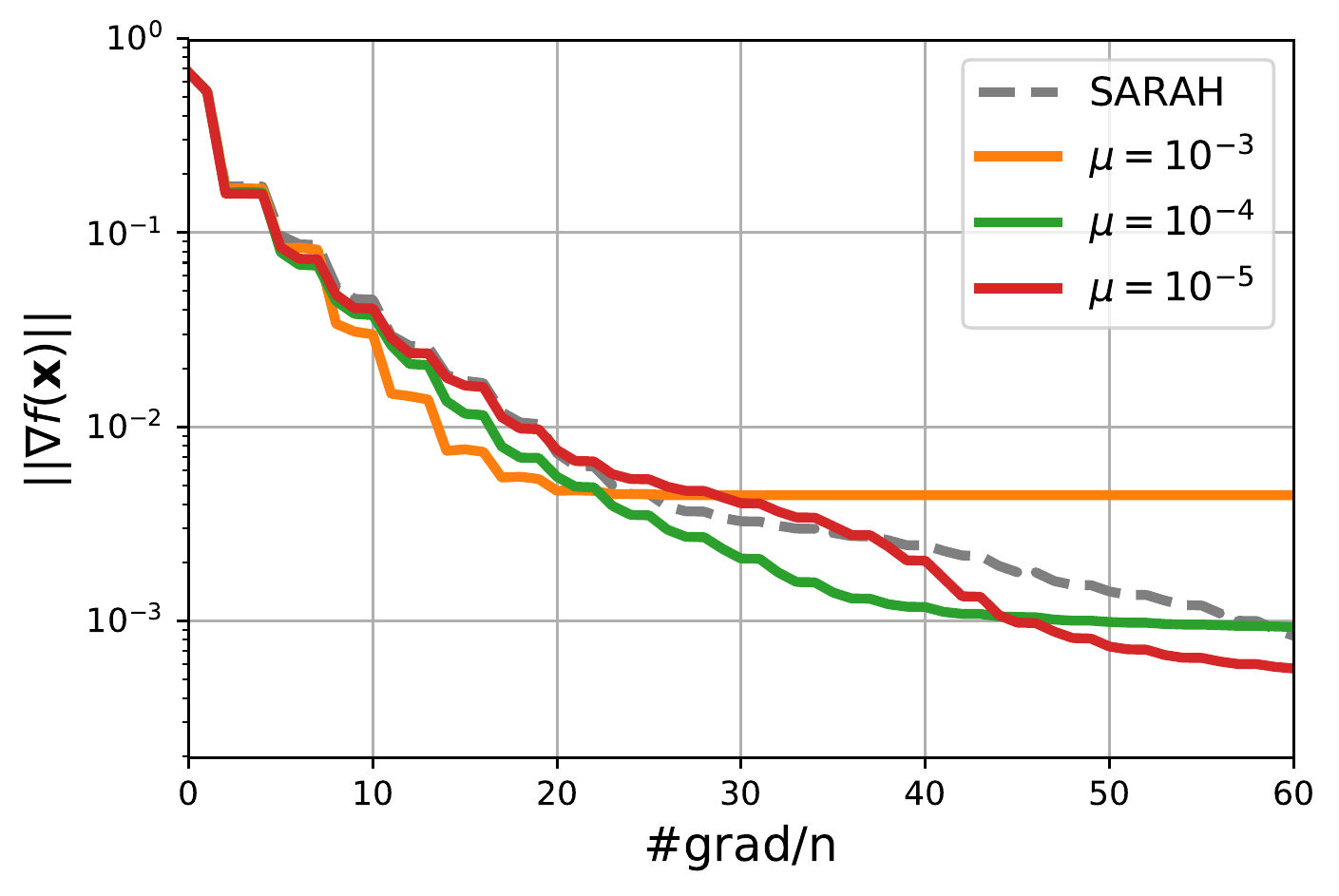}
	\caption{The ``plateau'' phenomenon in Alg. \ref{alg.single_reg} shown for a binary classification task on dataset \textit{a3a} using logistic regression.}  
	\label{fig.mu_choice}
%	\vspace{-0.2cm}
\end{tcolorbox}
\end{wrapfigure}

In this subsection, a \textit{recursive} regularization technique is developed for SARAH. Such regularization is motivated by the practical performance of Alg. \ref{alg.single_reg}. As we can see in Fig. \ref{fig.mu_choice}, the choice of $\mu$ is critical. In the experiment we observe that i) Alg. \ref{alg.single_reg} converges more rapidly for a large $\mu$ (e.g. $\mu=10^{-3}$) at the beginning stage; ii) there exists a ``platform'', that is, after certain iterations, $\| \nabla f(\mathbf{x}) \|$ almost remains unchange; iii) the larger $\mu$ is, the earlier the platform appears. The platform phenomenon happens when the surrogate function $\tilde{f}(\mathbf{x})$ is almost optimized. In general the choice of $\mu$ encounters dilemma: a large $\mu$ leads to faster convergence initially but early ``platform''; while small $\mu$ slows the convergence, but later ``platform''.

This subsection introduces a \textit{recursive} regularization technique motivated by the practical performance of Alg. \ref{alg.single_reg}. As evidenced by Fig. \ref{fig.mu_choice} (notice that the y-axis is log-scaled), the choice of $\mu$ is critical. Indeed, we observe experimentally that: i) Alg. \ref{alg.single_reg} converges more rapidly for a large $\mu$ (e.g. $\mu=10^{-3}$) at the beginning; ii) a ``plateau'' emerges afterwards; that is, $\| \nabla f(\mathbf{x}) \|$ almost remains unchanged  after certain iterations; iii) as $\mu$ grows larger, the plateau appears earlier. The plateau emerges when the surrogate function $\tilde{f}(\mathbf{x})$ is approximately optimized. In general, one faces a dilemma in selecting $\mu$: a large $\mu$ leads to faster convergence initially but reaches a ``plateau'' early; while a small $\mu$ slows the convergence down, but the plateau emerges later. This motivates us to adopt recursive regularization, that is, adopting a large $\mu$ to obtain fast converge at early stage, then decreasing $\mu$ to mitigate the ``plateau''.

Bearing this in mind, consider the surrogate function sequence as 
\begin{align}\label{eq.rr_reg}
%	\tilde{f}_0(\mathbf{x}) &= f(\mathbf{x}); \nonumber \\
	\tilde{f}_s(\mathbf{x}) &= f(\mathbf{x}) + \frac{\mu_s}{2} \| \mathbf{x} - \mathbf{x}_0 \|^2, ~\forall s \geq 1
\end{align}
where $\mathbf{x}_0$ is the initialization, and $\{\mu_s\}$ is a sequence of decreasing positive numbers to be specified later. Note that our surrogate functions are different from that of \citep{allen2018,foster2019} in which the strongly convexity of surrogate functions increases with $s$.

\begin{theorem}\label{thm.rsarah1}
	With Assumptions \ref{as.1} and \ref{as.2} holding, Alg. \ref{alg.recursive_sarah} v1 guarantees the total complexity for finding $\mathbb{E}[\| \nabla f(\tilde{\mathbf{x}}_S) \|^2] \leq \epsilon$ is ${\cal O}\big( (n+\frac{1}{\sqrt{\epsilon}}) \big(S+\ln \frac{1}{\epsilon}\big)\big)$ if one chooses the parameters as $ {\cal O}(\sqrt{\epsilon}) =\mu_1 >\mu_2 > \ldots > \mu_S = {\cal O}(\sqrt{\epsilon})$, $\eta_s < \frac{1}{L+\mu_s}$, $m_s = {\cal O}(\frac{L+\mu_s}{\mu_s})$, $K_1 = {\cal O}(\ln \frac{\| \nabla f(\mathbf{x}_0) \|^2}{\epsilon})$, and $K_s = {\cal O}(1), \forall s>1$.
\end{theorem}
In Theorem \ref{thm.rsarah1}, we choose $K_1$ to be sufficiently large so that after finishing $\texttt{SARAH}(\tilde{f}_1, \mathbf{x}_0, \eta_1, m_1, K_1)$, $\tilde{\mathbf{x}}_1$ is close to $\mathbf{x}^*$. With $K_s = {\cal O}(1), \forall s>1$ proximity to $\mathbf{x}^*$ is guaranteed, while the decreasing sequence $\{\mu_s\}$ ensures to get closer to $\mathbf{x}^*$ as $s$ increases.

\subsubsection{SARAH with Recursive Regularizer v2}
Though the recursive regularization can improve the performance of single regularization as confirmed by simulations in Section \ref{sec.sim}, it does not solve the ``plateau'' phenomenon since $\mu_S = {\cal O}(\sqrt{\epsilon})$. Next, we will introduce another variant of recursive regularization technique whose merits are: i) enhanced robustness to the parameter selection, i.e., choosing $\mu_s = {\cal O}(\sqrt{\epsilon}), \forall s$ is not necessary; ii) elimination of the plateau. It worth mentioning that in the regularization technique in \citep{allen2018}, the plateau phenomenon on convex problems is inevitable since a small regularized term is added there to provide strongly convexity.

Here we introduce a new sequence of surrogate functions as (cf. \eqref{eq.rr_reg})
\begin{align}\label{eq.rr_reg2}
%	\tilde{f}_0(\mathbf{x}) &= f(\mathbf{x}); \nonumber \\
	 \tilde{f}_s(\mathbf{x})& = f(\mathbf{x}) + \frac{\mu_s}{2} \| \mathbf{x} - \tilde{\mathbf{x}}_{s-1} \|^2, \forall s \geq 1
\end{align}
where $\tilde{\mathbf{x}}_0 = \mathbf{x}_0$ is the initialization; $\tilde{\mathbf{x}}_s, \forall s\geq 1$ is an approximation of $\tilde{\mathbf{x}}^*_s := \argmin_{\mathbf{x}}\tilde{f}_s(\mathbf{x})$; and the choices of $\mu_s$ will be specified later. The surrogate functions in \eqref{eq.rr_reg2} are designed such that $\nabla \tilde{f}_s(\tilde{\mathbf{x}}_{s-1}) = \nabla f(\tilde{\mathbf{x}}_{s-1})$, which is the key for convergence analysis. Leveraging \eqref{eq.rr_reg2} the designed algorithm is summarized in Alg. \ref{alg.recursive_sarah}.

\begin{theorem}\label{thm.rsarah2}
Under Assumptions \ref{as.1} and \ref{as.2}, select $\eta_s < \frac{1}{L+\mu_s}$, $m_s = {\cal O}(\frac{L+\mu_s}{\mu_s})$, and $K_s = {\cal O}\big(\ln \frac{1}{\rho_s}\big)$ such that $\mathbb{E}[\| \nabla \tilde{f}_s(\tilde{\mathbf{x}}_s ) \|^2] \leq \rho_s \mathbb{E}[\| \nabla \tilde{f}_s(\tilde{\mathbf{x}}_{s-1} ) \|^2]$ after each call of $\texttt{SARAH}(\tilde{f}_s, \tilde{\mathbf{x}}_{s-1}, \eta_s, m_s, K_s)$, where $\rho_s<1$. Choosing $\mu_s = \mu_0 (\Pi_{\tau=1}^s \rho_\tau)$ with $\mu_0 = {\cal O}(1)$, Alg. \ref{alg.recursive_sarah} v2 guarantees that
	\begin{align}\label{eq.noplateu}
		 \mathbb{E}\big[ \| \nabla f(\tilde{\mathbf{x}}_s) \|^2 \big]	 \!\leq \Big[ \! \prod_{\tau=1}^s \rho_\tau  \! \Big] \| \nabla f(\mathbf{x}_0) \|^2 + \mu_0 \Big[\! \prod_{\tau=1}^s \rho_\tau \! \Big] \Big(\! f(\mathbf{x}_0) \!-\! f(\mathbf{x}^*)\! \Big), \forall s. 	\end{align}
	In addition, the complexity to find $\mathbb{E}[ \| \nabla f(\tilde{\mathbf{x}}_S) \|^2 ]\leq \epsilon$ is ${\cal O}\big( \sum_{s=1}^S \big[ \big(n+ \frac{L}{\mu_s} \big) \ln \frac{1}{\rho_s}  \big] \big)$, which is strictly less than ${\cal O}\big( (n+\frac{L}{\epsilon})\ln \frac{1}{\epsilon}\big)$.
\end{theorem}

\subsection{Nonconvex Case}
This subsection focuses on regularization schemes for nonconvex problems. To this end, we start with the following definition\footnote{Also known as $\sigma$-weakly convex.}.

%%%%%%%%%%%%%%%%%%%%%%%%%%%%%%%%%%%%%%%%%%%%%%%%%%%%%%%%%%%
\begin{algorithm}[t]
    \caption{Recursively Regularized SARAH (nonconvex)}\label{alg.rr_reg_ncvx}
    \begin{algorithmic}[1]
    	\State \textbf{Initialize:} $\mathbf{x}_0$
    	\For {$s=1,2,\dots,S$}
    		\State define $\tilde{f}_s$ as in \eqref{eq.ncvx_surogate}
			\State $\tilde{\mathbf{x}}_s \leftarrow \texttt{SARAH}(\tilde{f}_s, \tilde{\mathbf{x}}_{s-1}, \eta_s, m_s, K_s)$ 
		\EndFor
		\State choose $k$ uniformly at random from $\{1,\ldots,S \}$
		\State \textbf{Return:} $\tilde{\mathbf{x}}_k$
	\end{algorithmic}
\end{algorithm}
%%%%%%%%%%%%%%%%%%%%%%%%%%%%%%%%%%%%%%%%%%%%%%%%%%%%%%%%%%

\begin{definition}
	($\sigma$-bounded nonconvex.) A continuously differentiable function $f$ is $\sigma$-bounded nonconvex  on $\mathbb{R}^d$ if there exists $\sigma >0$ such that $
		f(\mathbf{y}) - f(\mathbf{x}) \geq \big\langle  \nabla f(\mathbf{x}), \mathbf{y} - \mathbf{x} \big\rangle	 - \frac{\sigma}{2} \| \mathbf{y} - \mathbf{x} \|^2, ~\forall \mathbf{x} \in \mathbb{R}^d, \mathbf{y}\in \mathbb{R}^d.$
\end{definition}
By definition, any $L$-smooth function is $L$-bounded nonconvex, which implies that $\sigma \leq L$. Aiming to turn a $\sigma$-bounded nonconvex $f$ into strongly convex ones, we introduce the following the surrogate functions
\begin{align}\label{eq.ncvx_surogate}
	\tilde{f}_s(\mathbf{x}) &= f(\mathbf{x}) +\Big( \frac{\sigma}{2}  + \frac{\theta}{2} \Big) \| \mathbf{x} - \tilde{\mathbf{x}}_{s-1} \|^2, \forall s\geq 1
\end{align}
where $\tilde{\mathbf{x}}_0 = \mathbf{x}_0$ is the initialization; and $\tilde{\mathbf{x}}_s, \forall s\geq 1$ is an approximation of $\tilde{\mathbf{x}}^*_s := \argmin_{\mathbf{x}}\tilde{f}_s(\mathbf{x})$. It can be readily verified that $\tilde{f}_s$ is $(L+\theta+\sigma)$-smooth and $\theta$-strongly convex. Upon recursively optimizing \eqref{eq.ncvx_surogate}, the convergence of on the \textit{nonconvex} problem can be obtained as well. The resultant algorithm is summarized in Alg. \ref{alg.rr_reg_ncvx}.

\begin{theorem}\label{thm.rsarah_ncvx}
Suppose that Assumption \ref{as.1} holds. Choose $\eta_s < \frac{1}{L+\sigma+\theta}$, $m_s = {\cal O}(\frac{L+\sigma+\theta}{\theta})$, and $K_s = {\cal O}\big(\ln \frac{1}{\rho_s}\big)$ such that $\mathbb{E}[\| \nabla \tilde{f}_s(\tilde{\mathbf{x}}_s ) \|^2] \leq \rho_s \mathbb{E}[\| \nabla \tilde{f}_s(\tilde{\mathbf{x}}_{s-1} ) \|^2]$ after each call of $\texttt{SARAH}(\tilde{f}_s, \tilde{\mathbf{x}}_{s-1}, \eta_s, m_s, K_s)$, where $\rho_s<1$. Define $\rho:= \max_s \rho_s$. Alg. \ref{alg.rr_reg_ncvx} guarantees that 	
	\begin{align*}
		\mathbb{E} \big[ \|  \nabla f(\tilde{\mathbf{x}}_k) \|^2 \big]  \leq \frac{\rho }{(1-\rho) S} \| \nabla f(\mathbf{x}_{0}) \|^2 + \frac{2 (\sigma + \theta)	  \big[ f(\mathbf{x}_0) - f(\mathbf{x}^*) \big]}{(1-\rho)S}.
	\end{align*}	
To find $\mathbb{E} [ \|  \nabla f(\tilde{\mathbf{x}}_k) \|^2 ]\leq \epsilon $, choosing $\rho$ properly and $S = {\cal O}(1/\epsilon)$, the complexity of Alg. \ref{alg.rr_reg_ncvx} is
\begin{align*}
	{\cal O}  \bigg(  \frac{1}{\epsilon(1-\rho)}\Big(n + \frac{L+\sigma+\theta}{\theta} \Big) \ln \frac{1}{\rho} \bigg).
\end{align*}
\end{theorem}

Though the complexity is worse than the ${\cal O}\big(n+\frac{\sqrt{n}}{\epsilon}\big)$ of vanilla SARAH for nonconvex problems, the regularized SARAH is the foundation of the BB step size. It is also worth mentioning that in vanilla SARAH, the step size is often chosen as ${\cal O}\big( \frac{1}{\sqrt{n}L} \big)$, while due to the strongly convexity of $\tilde{f}_s$ in Alg. \ref{alg.rr_reg_ncvx}, a much larger step size ${\cal O}\big( \frac{1}{L+\sigma+\theta} \big)$ can be used. Meanwhile, the performance is robust to the choice of $\rho_s$ long as $\rho_s < 1, \forall s$.

\section{BB Step Size for (Non)Convex Problems}
In this section, BB step sizes for both convex and nonconvex problems are developed based on the regularization techniques.

\subsection{Convex Case}
As the surrogate functions in \eqref{eq.reg_f}, \eqref{eq.rr_reg} and \eqref{eq.rr_reg2} are already strongly convex, one can directly apply BB step size for minimizing them. For example, simply substituting line 3 of Alg. \ref{alg.single_reg} with $\tilde{\mathbf{x}} \leftarrow \texttt{BB-SARAH}(\tilde{f}, \mathbf{x}_{0}, \eta, m, K)$, one obtains the BB step size for the convex problem.
\begin{corollary}\label{coro.bbsarah_cvx}
(BB version of Alg. \ref{alg.single_reg}.) Choosing $\mu = {\cal O}\big(\frac{\sqrt{\epsilon}}{\|\mathbf{x}_0 - \mathbf{x}^* \|}\big)$, and applying BB-SARAH to solve $\min_{\mathbf{x}} \tilde{f}(\mathbf{x})$, the overall complexity for finding $ \mathbb{E} \big[ \| \nabla f(\mathbf{x}) \|^2 \big] \leq \epsilon$ is
	\begin{align*}
		{\cal O} \bigg( \Big(n+  \frac{L^2 \| \mathbf{x}_0 - \mathbf{x}^* \|^2}{ \epsilon} \Big) \ln \frac{\| \nabla f(\mathbf{x}_0) \|^2}{\epsilon} \bigg).
	\end{align*}
\end{corollary}

Corollary \ref{coro.bbsarah_cvx} suggests that equipping BB step size on SARAH will not introduce extra complexity compared with vanilla SARAH for convex problems \citep{nguyen2017}. This is surprising since even for strongly convex problems, the price of BB step size is a worse complexity bound. Another variant of BB step size was studied in \citep{ma2018}, however with a worse complexity ${\cal O}\big(n+ \frac{n}{\epsilon}\big)$; see Appendix \ref{apdx.ma2018} for details. In addition, two parameters should be tuned in \citep{ma2018}, while the BB-version of Alg. \ref{alg.single_reg} can almost eliminate tuning if one adopts ``tune-free'' BB-SARAH such as \citep{li2019bb} to minimize $\tilde{f}$.
 
Next we focus on the BB step size for recursively regularized SARAH, which is achieved by substituting line 5 of Alg. \ref{alg.recursive_sarah} with $\tilde{\mathbf{x}}_s \leftarrow \texttt{BB-SARAH}(\tilde{f}_s, \tilde{\mathbf{x}}_{s-1}, \eta_s, m_s, K_s)$.

\begin{corollary}
(BB version of Alg. \ref{alg.recursive_sarah} v1.) Choosing $ {\cal O}(\sqrt{\epsilon}) =\mu_1 >\mu_2 > \ldots > \mu_S = {\cal O}(\sqrt{\epsilon})$ and other parameters properly, the overall complexity for finding $ \mathbb{E} \big[ \| \nabla f(\mathbf{x}) \|^2 \big] \leq \epsilon$ is ${\cal O}\big( (n+\frac{1}{\epsilon}) (S+ \ln \frac{1}{\epsilon} ) \big)$, where ${\cal O}$ hides $L$ and constants. 
\end{corollary}
Corollary 2 suggests that recursive regularization v1 is free compared to vanilla SARAH on convex problems when choosing $S \leq {\cal O}\big(\ln \frac{1}{\epsilon}\big)$.

The BB step size for recursively regularized SARAH v2 can be obtained by substituting line 5 of Alg. \ref{alg.recursive_sarah} with $\tilde{\mathbf{x}}_s \leftarrow \texttt{BB-SARAH}(\tilde{f}_s, \tilde{\mathbf{x}}_{s-1}, \eta_s, m_s, K_s)$.
\begin{corollary}
(BB version of Alg. \ref{alg.recursive_sarah} v2.) Choosing $\{\mu_s\}$ and $\{\rho_s\}$ the same as Theorem \ref{thm.rsarah2}, the overall complexity for finding $ \mathbb{E} \big[ \| \nabla f(\mathbf{x}_S) \|^2 \big] \leq \epsilon$ is ${\cal O}\big( \sum_{s=1}^S \big[ \big(n+ \frac{L^2}{\mu_s^2} \big) \ln \frac{1}{\rho_s}  \big] \big)$. In addition, the complexity is strictly less than ${\cal O}\big( (n+\frac{L^2}{\epsilon^2})\ln \frac{1}{\epsilon}\big)$.
\end{corollary}
Corollary 3 suggests that the price paid to obtain robustness in parameter selection while avoiding the plateau phenomenon, is a worse $\epsilon$ dependence compared to vanilla SARAH. This worse $\epsilon$ dependence originates from the difficulty of using BB step size for optimizing strongly convex function with a large condition number. Hence, compared to the modified BB step size in \citep{ma2018}, when $n>\frac{1}{\epsilon}$ (in the big data regime or when a less accurate solution is desired), BB version of Alg. \ref{alg.recursive_sarah} v2 is preferable, otherwise one can rely on \citep{ma2018}.

\subsection{Nonconvex Case}
The BB version of SARAH for nonconvex problem can be obtained similarly by substituting line 4 of Alg. \ref{alg.rr_reg_ncvx} with $\tilde{\mathbf{x}}_s \leftarrow \texttt{BB-SARAH}(\tilde{f}_s, \tilde{\mathbf{x}}_{s-1}, \eta_s, m_s, K_s)$. 

\begin{corollary}
(BB version of Alg. \ref{alg.rr_reg_ncvx}.) Choose parameters properly such that $\mathbb{E}[\| \nabla \tilde{f}_s(\tilde{\mathbf{x}}_s ) \|^2] \leq \rho_s \mathbb{E}[\| \nabla \tilde{f}_s(\tilde{\mathbf{x}}_{s-1}) \|^2]$ for some $\rho_s < 1$. Denote $\rho:= \max_s \rho_s$, then the complexity for finding $\mathbb{E}[\| \nabla f(\mathbf{x}_k) \|^2 ]\leq \epsilon$ is ${\cal O}\big(\frac{n}{(1-\rho)\epsilon} \ln \frac{1}{\rho}\big)$.
\end{corollary}
	
The proposed BB version of Alg. \ref{alg.rr_reg_ncvx} has the same complexity as those in \citep{ma2018}, and both worse than vanilla SARAH on the $n$ dependence. This is the cost of being able to automatically tune the step size. Note that as the strongly convexity remains $\theta$ for all $s$ in \eqref{eq.ncvx_surogate}, the step size used in the last inner loop of minimizing $\tilde{f}_s$ is a valid for the step size initialization of $\tilde{f}_{s+1}$. Again, it is worth stressing that for being insensitive to the choice of$\rho_s$, the BB version of Alg. \ref{alg.rr_reg_ncvx} with the subproblems solved using \citep{li2019bb} is almost tune free.

Next we compare the proposed scheme to other adaptive step sizes such as AdaGrad \citep{duchi2011,mcmahan2010}, Adam \citep{kingma2014} and their variants \citep{reddi2019,ward2018,li2018convergence,zaheer2018}. For convenience we use AdaGrad to refer to all these algorithms. In theory, those comparisons are not necessarily fair since: i) AdaGrad is designed for stochastic optimization, i.e., $\min_{\mathbf{x}} \mathbb{E}_\xi[f(\mathbf{x},\xi)]$, while the proposed scheme focuses on finite sum problem \eqref{eq.prob}; and, ii) the assumptions made in analyses are different. For fairness, we first consider the ``deterministic'' setting, where AdaGrad calculate the full gradient $\nabla f(\mathbf{x}_k)$ per iteration and uses it for updating.

\textbf{Deterministic Setting.} Comparing AdaGrad in the deterministic setting with Alg. \ref{alg.rr_reg_ncvx} having BB step size is fair since: i) for AdaGrad to obtain $\nabla f(\mathbf{x}_k)$ amounts to reduce the stochastic optimization to finite sum problems; and, ii) the only assumption for both AdaGrad and proposed schemes is Assumption \ref{as.1}. The complexity of AdaGrad is ${\cal O}(\frac{n}{\epsilon})$, meaning identical to the proposed algorithm.

%Algorithms such as AdaGrad is more general than proposed scheme since in theory they can deal with stochastic optimization; while on the other hand, even for finite sum problems \eqref{eq.prob} most of analysis require at least one of the following extra assumptions i) bounded gradient, that is, $\mathbb{E}_i[\|\nabla f_{i} (\mathbf{x})\|^2] \leq {\cal G}$; ii) bounded variance  $\mathbb{E}_i[\|\nabla f_{i} (\mathbf{x}) - \nabla f(\mathbf{x})\|^2] \leq {\cal V}$.

\textbf{Stochastic Setting.} The stochastic setting is in general an unfair setting for comparison. Here we only consider using AdaGrad for solving \eqref{eq.prob}. Even for \eqref{eq.prob}, most of analyses of AdaGrad require at least one of the following extra assumptions i) bounded gradient, that is, $\mathbb{E}_i[\|\nabla f_{i} (\mathbf{x})\|^2] \leq {\cal G}$; ii) bounded variance  $\mathbb{E}_i[\|\nabla f_{i} (\mathbf{x}) - \nabla f(\mathbf{x})\|^2] \leq {\cal V}$. Regardless of these differences on assumptions compared with Alg. \ref{alg.rr_reg_ncvx}, the complexity of AdaGrad to find $\mathbb{E}[\| \nabla f(\mathbf{x}) \|^2] \leq \epsilon$ is ${\cal O} (\frac{ {\cal C} ({\cal G}, {\cal V} ) }{\epsilon^2})$, where ${\cal C} ({\cal G}, {\cal V} )$ is a constant depending on ${\cal G}$ and ${\cal V}$. Hence, when requiring an accurate solution or $n$ is small, Alg. \ref{alg.rr_reg_ncvx} with BB step size is preferable compared with AdaGrad. Notice that in the complexity bound of AdaGrad, both ${\cal G}$ and ${\cal V}$ implicitly depend on the dimensionality $d$; however, the complexity of Alg. \ref{alg.rr_reg_ncvx} with BB step size is dimensionality free.

\section{Implementation}\label{sec.sim}
This section deals with implementation and numerical tests of proposed approaches.

\begin{figure}[t]
	\centering
	\begin{tabular}{cc}
		\hspace{-0.4cm}
		\includegraphics[width=.45\textwidth]{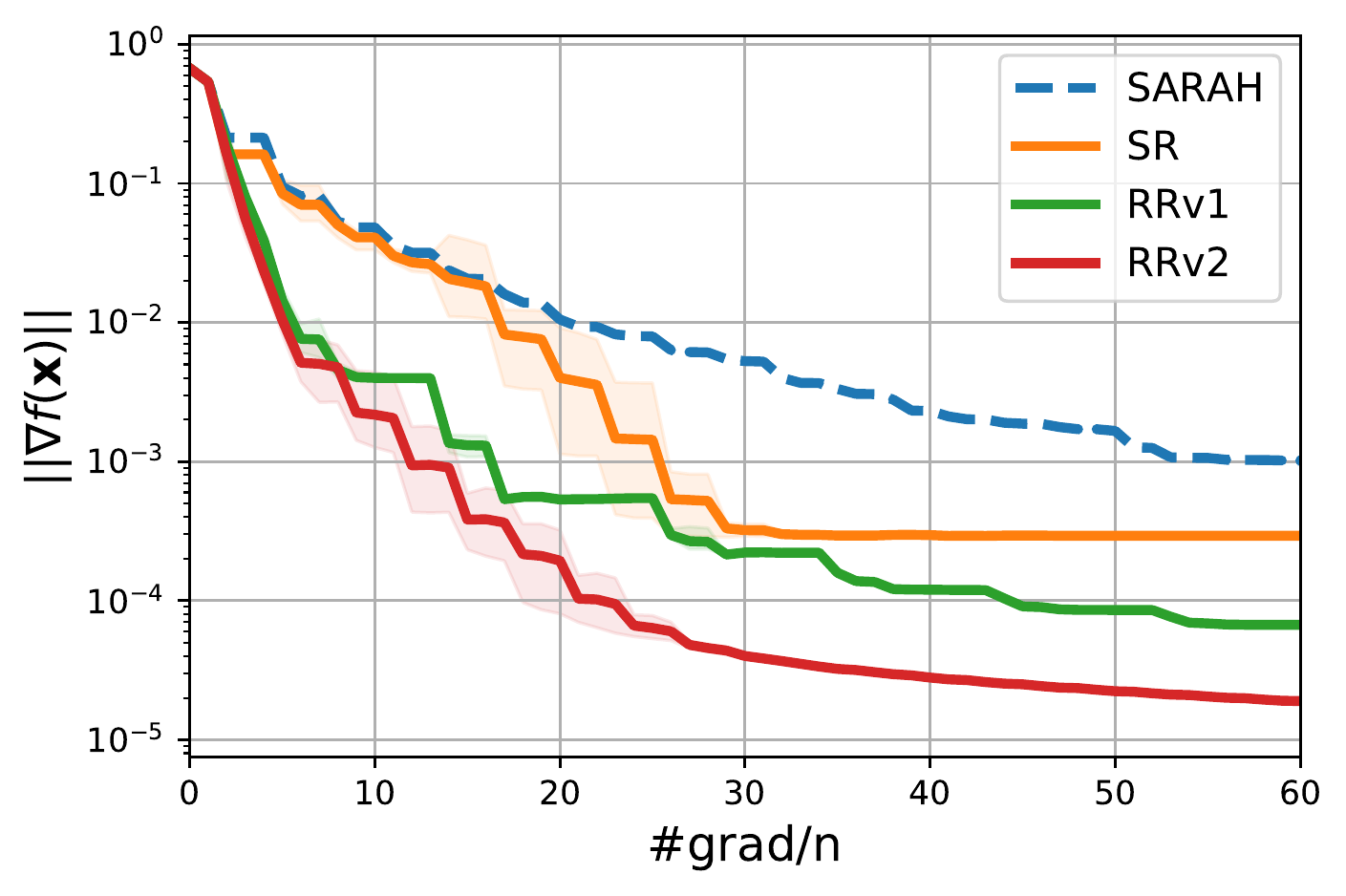}&
		\hspace{-0.4cm}
		\includegraphics[width=.45\textwidth]{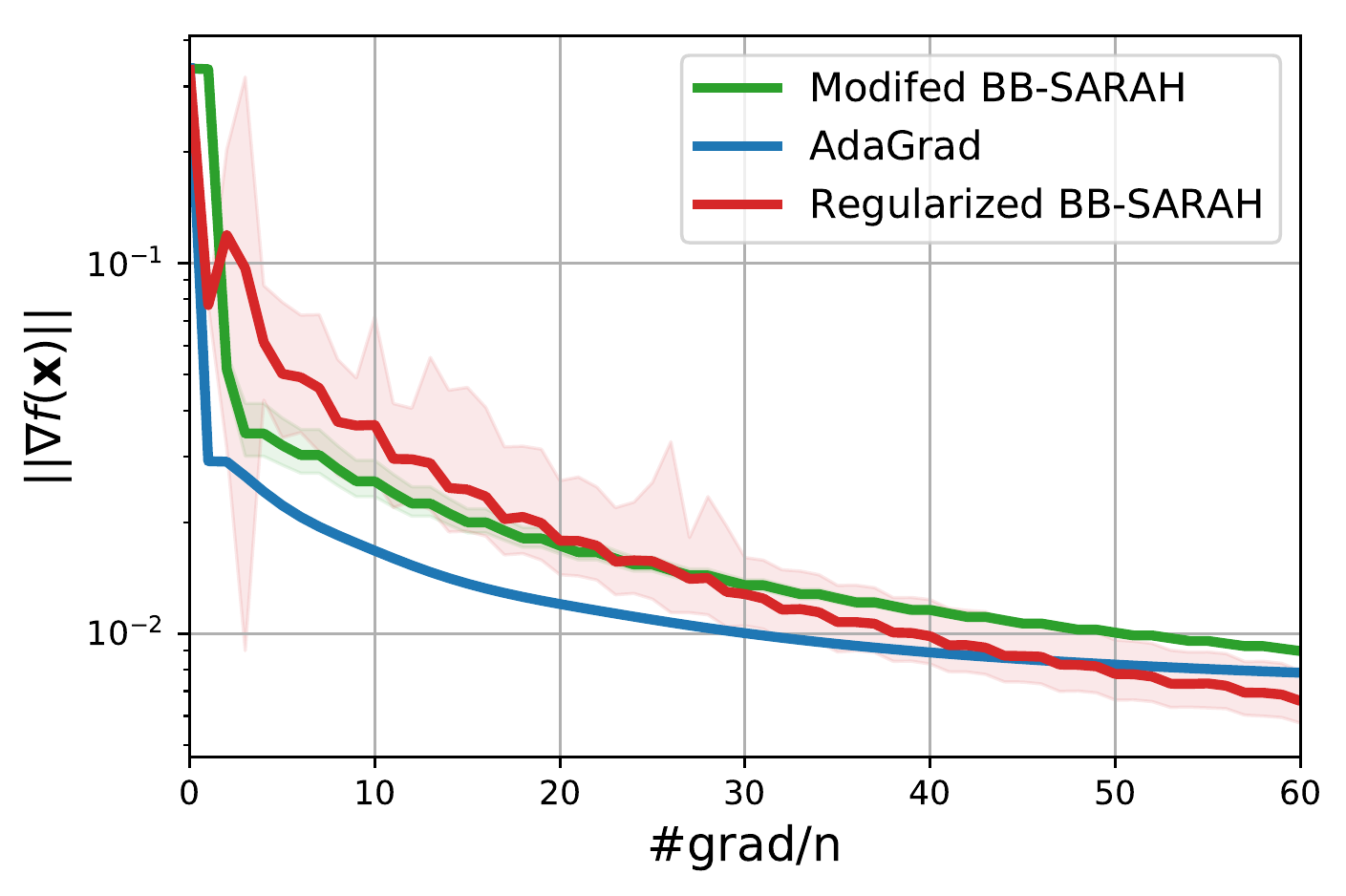}
		\\ (a) convex  & (b) nonconvex 
	\end{tabular}
	\caption{Numerical tests of proposed approaches on dataset \textit{a3a}.}
	\label{fig.tests}
\end{figure}

\subsection{Implementation Tricks}
We first introduce tricks for efficient implementation of our recursively regularized SARAH. We will use Alg. \ref{alg.recursive_sarah} v2 as an example.

It holds because of \eqref{eq.rr_reg2} that $\nabla \tilde{f}_s(\mathbf{x}) = \nabla f(\mathbf{x}) + \mu_s (\mathbf{x} - \tilde{\mathbf{x}}_{s-1})$. This equation suggests that whenever a full gradient of $\tilde{f}_s$ is computed, one can automatically obtain the gradient of objective function $f$. Hence it is possible to directly supervise the progress of optimization process.

In addition, instead of choosing $\rho_s$ in Theorem \ref{thm.rsarah2}, a simpler implementation alternative is to choose $K_s$ and then find $\rho_s = \|\nabla \tilde{f}_s(\tilde{\mathbf{x}}_s)\|^2 /\|\nabla \tilde{f}_s (\tilde{\mathbf{x}}_{s-1})\|^2 $. In this case, one can obtain a full gradient for $\tilde{f}_{s+1}(\tilde{\mathbf{x}}_s)$ as $\nabla \tilde{f}_{s+1}(\tilde{\mathbf{x}}_s)  = \nabla \tilde{f}_{s}(\tilde{\mathbf{x}}_s) - \mu_s (\tilde{\mathbf{x}}_s - \tilde{\mathbf{x}}_{s-1})$, so that there is no need to compute $\nabla \tilde{f}_{s+1}(\tilde{\mathbf{x}}_s)$ again when minimizing $\tilde{f}_{s+1}$.

%This is obtained by combining $\nabla \tilde{f}_{s+1}(\tilde{\mathbf{x}}_s) = \nabla f (\tilde{\mathbf{x}}_s)$ and $\nabla \tilde{f}_s(\tilde{\mathbf{x}}_s) = \nabla f (\tilde{\mathbf{x}}_s) + \mu_s (\tilde{\mathbf{x}}_s - \tilde{\mathbf{x}}_{s-1})$.

\subsection{Numerical Tests} 
Simulations are carried out to validate the proposed schemes starting with convex case.

\textbf{Convex Case.} Consider binary classification with logistic regression
\begin{equation*}
	f(\mathbf{x}) =\frac{1}{n} \sum_{i \in [n]} \ln \big[1+ \exp(- b_i \langle \mathbf{a}_i, \mathbf{x} \rangle ) \big] 
\end{equation*}
where $\mathbf{a}_i$ ($b_i$) is the feature (label) of datum $i$. Performance
comparison of the proposed regularized SARAH against the vanilla SARAH are plotted in Fig. \ref{fig.tests} (a) for dataset \textit{a3a} from LIBSVM\footnote{Online available at \url{https://www.csie.ntu.edu.tw/~cjlin/libsvmtools/datasets/binary.html}}. Clearly, all proposed regularization schemes significantly improve over vanilla SARAH, while Alg. \ref{alg.recursive_sarah} v2 exhibits the best performance.

\textbf{Nonconvex Case.} Binary classification with the following nonconvex loss \citep{li2003} on dataset \textit{a3a} is considered
\begin{align*}
	f(\mathbf{x}) = \frac{1}{n}\sum_{i\in [n]} \bigg(1 - \frac{1}{1+\exp(-b_i \langle \mathbf{a}_i, \mathbf{x} \rangle)} \bigg)^2.
\end{align*}
The performances of Alg. \ref{alg.rr_reg_ncvx} with BB-SARAH for surrogate functions, the modified BB step size in \citep{ma2018} combined with SARAH, and AdaGrad \citep{duchi2011} are showcased in Fig. \ref{fig.tests} (b). AdaGrad and modified BB-SARAH both converge rapidly early on, but they are outperformed by the proposed method at final stage.

\section{Conclusions}
Barzilai-Borwein (BB) step size was broadened to encompass convex and nonconvex optimization problems with a finite-sum structure. The main challenge is the strongly convexity required to define an appropriate BB step size. This was successfully addressed through (recursively) adding a quadratic regularizer on the objective function to obtain a strongly convex surrogate, and then using BB step size to optimize the surrogate. Complexity of the proposed schemes was quantified, and validated through numerical tests.

\bibliographystyle{plainnat}
\bibliography{myabrv,datactr}

\newpage
\onecolumn
\appendix
\begin{center}
{\large  \bf Supplementary Document for ``Almost Tune-Free Variance Reduction'' }
\end{center}

\section{Facts of Strongly Convex Function}\label{sec.stronglycvx}

The required inequalities for proofs are summarized below.
\begin{lemma}\label{lemma.sc}
\citep[Theorems 2.1.7 and 2.1.9 ]{nesterov2004}
	If $f$ is $\mu$-strongly convex, then we have
	\begin{align}
		f(\mathbf{x}) & \geq f(\mathbf{x}^*) + \frac{\mu}{2} \| \mathbf{x} - \mathbf{x}^* \|^2  \label{eq.sc.3} \\
		 \mu \| \mathbf{x} - \mathbf{y} \|^2 & \leq  \big\langle \nabla f(\mathbf{x}) - \nabla f(\mathbf{y}), \mathbf{x} - \mathbf{y} \big\rangle \label{eq.sc.1}
	\end{align}
\end{lemma}

The consequence of Lemma \ref{lemma.sc} follows. By \eqref{eq.sc.1} and Cauchy-Schwarz inequality we have $\mu \| \mathbf{x} - \mathbf{y} \|^2 \leq  \big\langle \nabla f(\mathbf{x}) - \nabla f(\mathbf{y}), \mathbf{x} - \mathbf{y} \big\rangle \leq \|\nabla f(\mathbf{x}) - \nabla f(\mathbf{y}) \| \|\mathbf{x} - \mathbf{y} \|$. Cancelling $ \|\mathbf{x} - \mathbf{y} \|$ out, we arrive at
\begin{align}\label{eq.sc.2}
	\mu \| \mathbf{x} - \mathbf{y} \| \leq \|\nabla f(\mathbf{x}) - \nabla f(\mathbf{y}) \|.
\end{align}

%%%%%%%%%%%%%%%%%%%%%%%%%%%%%%%%%%%%%%%%%%%%%%%%%%%%%%%%%%%%%%%%%%%%%%%%%%%%%%%%%%%%%%%%%%%%%%%%%%%%%%%%%%%%%%%%%%%%%%%%%
\section{Missing Proofs}
\subsection{Proof of Theorem \ref{thm.single_reg}}
\begin{proof}
	Define $\mathbf{x}^* \in \argmin_{\mathbf{x}} f(\mathbf{x})$ and $\tilde{\mathbf{x}}^* = \argmin_{\mathbf{x}} \tilde{f}(\mathbf{x})$ as optimal solutions of $f$ and $\tilde{f}$, respectively. Using the definition of $\tilde{f}$, we then have
	\begin{align*}
		\nabla \tilde{f}(\mathbf{x}) =  \nabla f(\mathbf{x}) + \mu (\mathbf{x}-\mathbf{x}_0).
	\end{align*}
	
	The choices of $\eta$, $m$ and $K$ guarantee that after calling $\texttt{SARAH}(\tilde{f}, \tilde{\mathbf{x}}_{0}, \eta, m, K)$, a solution, $\mathbf{x}_a$, with $\mathbb{E}\big[\|\nabla \tilde{f}(\mathbf{x}_a) \|^2 \big] \leq \epsilon$ is obtained. Next we relate $\|\nabla \tilde{f}(\mathbf{x}_a) \|^2 $ and $\|\nabla f(\mathbf{x}_a) \|^2 $. To start, consider that
	\begin{align}\label{eq.c1}
		&~~~~~ \| \nabla f(\mathbf{x}_a) \|^2  = \| \nabla \tilde{f}(\mathbf{x}_a) - \mu (\mathbf{x}_a -\mathbf{x}_0) \|^2 \nonumber \\
		& = \| \nabla \tilde{f}(\mathbf{x}_a) - \mu (\mathbf{x}_a - \tilde{\mathbf{x}}^* + \tilde{\mathbf{x}}^* -\mathbf{x}_0) \|^2 \nonumber \\
		& \stackrel{(a)}{\leq} 3 \| \nabla \tilde{f}(\mathbf{x}_a) \|^2 + 3\mu^2 \| \mathbf{x}_a -\tilde{\mathbf{x}}^* \|^2 + 3\mu^2 \| \tilde{\mathbf{x}}^* -\mathbf{x}_0 \|^2  \nonumber \\
		&\stackrel{(b)}{\leq } 6 \| \nabla \tilde{f}(\mathbf{x}_a) \|^2 +  3\mu^2 \| \tilde{\mathbf{x}}^* -\mathbf{x}_0 \|^2
	\end{align}
	where (a) is because $\| \mathbf{a} + \mathbf{b} + \mathbf{c} \|^2 \leq 3 \| \mathbf{a} \|^2 + 3 \| \mathbf{b} \|^2 + 3 \| \mathbf{c} \|^2$; (b) follows from \eqref{eq.sc.2}, that is, $\| \nabla \tilde{f}(\mathbf{x}_a) \| = \| \nabla \tilde{f}(\mathbf{x}_a) -  \nabla \tilde{f}(\tilde{\mathbf{x}}^*) \| \geq \mu \| \mathbf{x}_a -  \tilde{\mathbf{x}}^* \|$. We then have 
	\begin{align}\label{eq.c2}
		\| \tilde{\mathbf{x}}^* -\mathbf{x}_0 \|^2	& \leq 2 \| \tilde{\mathbf{x}}^* - \mathbf{x}^* \|^2 + 2 \|  \mathbf{x}^* -\mathbf{x}_0 \|^2 \nonumber \\
		&  \stackrel{(c)}{\leq} \frac{4}{\mu} \big[ \tilde{f}(\mathbf{x}^*) - \tilde{f}(\tilde{\mathbf{x}}^*) \big] + 2 \|  \mathbf{x}^* -\mathbf{x}_0 \|^2  \nonumber \\
		& \stackrel{(d)}{=} \frac{4}{\mu} \Big[ f(\mathbf{x}^*) + \frac{\mu}{2} \| \mathbf{x}^* - \mathbf{x}_0 \|^2\Big]  - \frac{4}{\mu}\Big[ f(\tilde{\mathbf{x}}^*) + \frac{\mu}{2} \| \tilde{\mathbf{x}}^* - \mathbf{x}_0 \|^2 \Big] + 2 \|  \mathbf{x}^* -\mathbf{x}_0 \|^2 \nonumber \\
		& \stackrel{(e)}{\leq} \frac{4}{\mu} \Big[  \frac{\mu}{2} \| \mathbf{x}^* - \mathbf{x}_0 \|^2  -\frac{\mu}{2} \| \tilde{\mathbf{x}}^* - \mathbf{x}_0 \|^2 \Big] + 2 \|  \mathbf{x}^* -\mathbf{x}_0 \|^2 \nonumber \\
		& \leq 4 \|  \mathbf{x}^* -\mathbf{x}_0 \|^2 
	\end{align}
	where (c) follows from \eqref{eq.sc.3}, that is, $ \| \mathbf{x} - \tilde{\mathbf{x}}^* \|^2 \leq \frac{2}{\mu} \big(  \tilde{f}(\mathbf{x}) - \tilde{f}(\tilde{\mathbf{x}}^*) \big), \forall \mathbf{x}$; (d) is obtained by using the definition of $\tilde{f}$; (e) is due to $ f(\mathbf{x}^*) - f(\tilde{\mathbf{x}}^*) \leq 0$. Plugging \eqref{eq.c2} into \eqref{eq.c1}, we arrive at
	\begin{align}\label{eq.c3}
		\| \nabla f(\mathbf{x}_a) \|^2 	\leq 6 \| \nabla \tilde{f}(\mathbf{x}_a) \|^2 +  12 \mu^2 \| \mathbf{x}_0 -  \mathbf{x}^*  \|^2 
	\end{align}
	
	Hence, so long as we choose $\mu = {\cal O}\big(\frac{\sqrt{\epsilon}}{\|\mathbf{x}_0 - \mathbf{x}^* \|}\big)$, having $\mathbb{E}[\| \nabla \tilde{f}(\mathbf{x}_a) \|^2] \leq \epsilon $ directly implies $\mathbb{E}[\| \nabla f(\mathbf{x}_a) \|^2] 	= {\cal O}(\epsilon)$. Such choice of $\mu$ leads to the condition number of $\tilde{f}$ being $\tilde{\kappa} = {\cal O} \big( 1 + \frac{L \| \mathbf{x}_0 - \mathbf{x}^* \|}{ \sqrt{\epsilon}} \big) = {\cal O} \big( \frac{L \| \mathbf{x}_0 - \mathbf{x}^* \|}{ \sqrt{\epsilon}} \big)$.
	
	It is well known the complexity of SARAH to solve $\min_{\mathbf{x}} \tilde{f}(\mathbf{x})$ with guaranteed $\mathbb{E}[\| \nabla \tilde{f} (\mathbf{x}) \|^2]\leq \epsilon$ is ${\cal O}\big( (n+\tilde{\kappa}) \ln \frac{\| \nabla \tilde{f}(\mathbf{x}_0) \|^2}{\epsilon}\big) = {\cal O}\big( (n+\tilde{\kappa}) \ln \frac{\| \nabla f(\mathbf{x}_0) \|^2}{\epsilon}\big)$ \citep{nguyen2017}. Substituting $\tilde{\kappa}$ in the latter, we can quantify the complexity of Alg. \ref{alg.single_reg}	as 
	\begin{align*}
		{\cal O} \bigg( \Big(n+  \frac{L \| \mathbf{x}_0 - \mathbf{x}^* \|}{ \sqrt{\epsilon}} \Big) \ln \frac{\| \nabla f(\mathbf{x}_0) \|^2}{\epsilon} \bigg)
	\end{align*}
	which completes the proof.
\end{proof}

%%%%%%%%%%%%%%%%%%%%%%%%%%%%%%%%%%%%%%%%%%%%%%%%%%%%%%%%%%%%%%%%%%%%%%%%%%%%%%%%%%%%%%%%%%%%%%%%%%%%%%%%%%%%%%%%%%%%%%%%%
\subsection{Proof of Theorem \ref{thm.rsarah1}}
Suppose that after $s$-th call of SARAH, we obtain $\tilde{\mathbf{x}}_s$, and let $\tilde{\mathbf{x}}_s^*:= \argmin_{\mathbf{x}} \tilde{f}_s(\mathbf{x})$. Using similar arguments as Theorem \ref{thm.single_reg} and definition of $\tilde{f}_s$ in \eqref{eq.rr_reg}, we have
\begin{align*}
	\| \nabla f(\tilde{\mathbf{x}}_s) \|^2 	\leq 6 \| \nabla \tilde{f}_s(\tilde{\mathbf{x}}_s) \|^2 +  12 \mu_s^2 \| \mathbf{x}_0 -  \mathbf{x}^*  \|^2, \forall s.
\end{align*}
This means that long as we choose $\mu_s = {\cal O}(\sqrt{\epsilon}), \forall s$, and each time minimize $\tilde{f}_s$ such that  $\mathbb{E}[ \| \nabla \tilde{f}_s (\tilde{\mathbf{x}}_s) \|^2 ] \leq \epsilon$, we are always guaranteed to have $\mathbb{E}[\| \nabla f(\tilde{\mathbf{x}}_s)\|^2 ] = {\cal O}(\epsilon)$. Hence, the choices of $\mu_s = {\cal O}(\sqrt{\epsilon}), \forall s$ guarantees $\mathbb{E}[\| \nabla f(\tilde{\mathbf{x}}_s) \|^2 ]={\cal O}(\epsilon) ,\forall s$ if $\mathbb{E} [ \| \nabla \tilde{f}_s(\tilde{\mathbf{x}}_s) \|^2] = {\cal O}(\epsilon), \forall s$ is true.

To have $\mathbb{E} [ \| \nabla \tilde{f}_1(\tilde{\mathbf{x}}_1) \|^2] = {\cal O}(\epsilon)$, we have to choose $K_1 = {\cal O}(\ln \frac{\| \nabla f(\mathbf{x}_0)\|^2}{\epsilon})$. Then we will design $K_s$, so that the $\mathbb{E} [ \| \nabla \tilde{f}_s(\tilde{\mathbf{x}}_s) \|^2] = {\cal O}(\epsilon), \forall s \geq 1$ is satisfied. Consider that
\begin{align}
	\|  \nabla \tilde{f}_{s+1}(\tilde{\mathbf{x}}_s) \|^2 & = \big\|  \nabla f(\tilde{\mathbf{x}}_s) + \mu_{s+1}(\tilde{\mathbf{x}}_s - \mathbf{x}_0) \big\|^2 = \big\|  \nabla f(\tilde{\mathbf{x}}_s) + \mu_s (\tilde{\mathbf{x}}_s - \mathbf{x}_0) + ( \mu_{s+1} - \mu_s)(\tilde{\mathbf{x}}_s - \mathbf{x}_0) \big\|^2 \nonumber \\
	& = \big\|  \nabla \tilde{f}_s(\tilde{\mathbf{x}}_s) + ( \mu_{s+1} - \mu_s)(\tilde{\mathbf{x}}_s - \mathbf{x}_0) \big\|^2  = \big\|  \nabla \tilde{f}_s(\tilde{\mathbf{x}}_s) + ( \mu_{s+1} - \mu_s)(\tilde{\mathbf{x}}_s - \tilde{\mathbf{x}}_s^* + \tilde{\mathbf{x}}_s^*  - \mathbf{x}_0) \big\|^2  \nonumber \\
	& \stackrel{(a)}{\leq} 3 \big\|  \nabla \tilde{f}_s(\tilde{\mathbf{x}}_s)\big\|^2 + 3(\mu_s - \mu_{s+1})^2 \big\|  \tilde{\mathbf{x}}_s - \tilde{\mathbf{x}}_s^*  \big\|^2 + 3(\mu_s - \mu_{s+1})^2 \big\|   \tilde{\mathbf{x}}_s^*  - \mathbf{x}_0  \big\|^2 \nonumber \\
	& \stackrel{(b)}{\leq} 3 \big\|  \nabla \tilde{f}_s(\tilde{\mathbf{x}}_s)\big\|^2 + \frac{3(\mu_s - \mu_{s+1})^2}{\mu_s^2} \big\|  \nabla \tilde{f}_s(\tilde{\mathbf{x}}_s)\big\|^2 + 3(\mu_s - \mu_{s+1})^2 \big\|   \tilde{\mathbf{x}}_s^*  - \mathbf{x}_0  \big\|^2 \nonumber \\
	&\stackrel{(c)}{\leq} 3 \big\|  \nabla \tilde{f}_s(\tilde{\mathbf{x}}_s)\big\|^2 + \frac{3(\mu_s - \mu_{s+1})^2}{\mu_s^2} \big\|  \nabla \tilde{f}_s(\tilde{\mathbf{x}}_s)\big\|^2 + 12 (\mu_s - \mu_{s+1})^2 \big\|   \mathbf{x}^*  - \mathbf{x}_0  \big\|^2 \nonumber
\end{align}
where (a) is because $\| \mathbf{a} + \mathbf{b} + \mathbf{c} \|^2 \leq 3 \| \mathbf{a} \|^2 + 3 \| \mathbf{b} \|^2 + 3 \| \mathbf{c} \|^2$; (b) is by strongly convexity of $\tilde{f}_s$ and \eqref{eq.sc.2}; and in (c) we use similar argument as \eqref{eq.c2} in Theorem \ref{thm.single_reg}.

Therefore, choosing $\mu_{s+1} \leq \mu_s = {\cal O}(\sqrt{\epsilon})$ guarantees when $\big\|  \nabla \tilde{f}_s(\tilde{\mathbf{x}}_s)\big\|^2 ={\cal O} (\epsilon)$ holds, $\|  \nabla \tilde{f}_{s+1}(\tilde{\mathbf{x}}_s) \|^2 = {\cal O}(\epsilon)$ as well. Hence to guarantee $\|  \nabla \tilde{f}_{s+1}(\tilde{\mathbf{x}}_{s+1}) \|^2 = {\cal O}(\epsilon)$, one only need to choose $K_s = {\cal O}(1)$ with proper $\eta_s$ and $m_s$ such that SARAH is converging.

The overall complexity hence becomes
\begin{align*}
	\sum_{s=1}^S {\cal O}\bigg( \Big(n+ \frac{L+\mu_s}{\mu_s} \Big)K_s \bigg) & = {\cal O}\bigg( \Big(n+ \frac{L+\mu_1}{\mu_1} \Big) K_1 \bigg)  + \sum_{s=2}^S {\cal O} \bigg( \Big(n+ \frac{L+\mu_s}{\mu_s} \Big)K_s \bigg) \nonumber \\
	& \leq {\cal O}\bigg( \Big(n+ \frac{L}{\sqrt{\epsilon}} \Big) \bigg( S+ \ln \frac{1}{\epsilon} \bigg) \bigg).
\end{align*}

%%%%%%%%%%%%%%%%%%%%%%%%%%%%%%%%%%%%%%%%%%%%%%%%%%%%%%%%%%%%%%%%%%%%%%%%%%%%%%%%%%%%%%%%%%%%%%%%%%%%%%%%%%%%%%%%%%%%%%%%%
\subsection{Proof of Theorem \ref{thm.rsarah2}}
\begin{proof}
	By the definition of $\tilde{f}_s$, we have $\nabla \tilde{f}_s(\mathbf{x}) = \nabla f(\mathbf{x}) + \mu_s (\mathbf{x} - \tilde{\mathbf{x}}_{s-1})$, which implies that
	\begin{align*}
		  \| \nabla \tilde{f}_s(\tilde{\mathbf{x}}_s) \|^2  & =\! \|  \nabla f(\tilde{\mathbf{x}}_s) \|^2  \!+\! \mu_s^2 \| \tilde{\mathbf{x}}_s -\tilde{\mathbf{x}}_{s\!-\!1} \|^2 \!+\! 2 \mu_s \big\langle \nabla f(\tilde{\mathbf{x}}_s) , \tilde{\mathbf{x}}_s -\tilde{\mathbf{x}}_{s\!-\!1} \big\rangle \nonumber \\
		 &\stackrel{(a)}{\geq} \|  \nabla f(\tilde{\mathbf{x}}_s) \|^2 + \mu_s^2 \| \tilde{\mathbf{x}}_s -\tilde{\mathbf{x}}_{s\!-\!1} \|^2 + 2\mu_s \big[ f(\tilde{\mathbf{x}}_s) - f(\tilde{\mathbf{x}}_{s\!-\!1}) \big]
	\end{align*}
	where (a) follows from the convexity of $f$. Rearranging the terms and taking expectation leads to
	\begin{align*}
		 \mathbb{E} \big[ \|  \nabla f(\tilde{\mathbf{x}}_s) \|^2 \big] & \leq  	\mathbb{E} \big[ \| \nabla \tilde{f}_s(\tilde{\mathbf{x}}_s) \|^2 \big] - \mu_s^2 	\mathbb{E} \big[ \| \tilde{\mathbf{x}}_s -\tilde{\mathbf{x}}_{s-1} \|^2\big]  + 2\mu_s 	\mathbb{E}\big[ f(\tilde{\mathbf{x}}_{s-1}) - f(\tilde{\mathbf{x}}_s) \big]  \\
		&\stackrel{(b)}{\leq} \rho_s \mathbb{E} \big[\| \nabla \tilde{f}_s(\tilde{\mathbf{x}}_{s-1}) \|^2\big] + 2\mu_s \mathbb{E}\big[ f(\tilde{\mathbf{x}}_{s-1}) - f(\tilde{\mathbf{x}}_s) \big] \\
		& \stackrel{(c)}{=} \rho_s 	\mathbb{E} \big[\| \nabla f(\tilde{\mathbf{x}}_{s-1}) \|^2 \big] + 2\mu_s \mathbb{E}\big[ f(\tilde{\mathbf{x}}_{s-1}) - f(\tilde{\mathbf{x}}_s) \big]
	\end{align*}
	where in (b) we use $\mathbb{E}[\| \nabla \tilde{f}_s(\tilde{\mathbf{x}}_s ) \|^2] \leq \! \rho_s \mathbb{E}[\| \nabla \tilde{f}_s(\tilde{\mathbf{x}}_{s\!-\!1} ) \|^2]$; and (c) is by the fact $\nabla f(\tilde{\mathbf{x}}_{s-1}) = \nabla \tilde{f}_s(\tilde{\mathbf{x}}_{s-1})$. Now we can unroll $\mathbb{E} \big[\| \nabla f(\tilde{\mathbf{x}}_{s-1}) \|^2 \big]$ to obtain
	\begin{align*}
		 \mathbb{E} \big[ \|  \nabla f(\tilde{\mathbf{x}}_s) \|^2 \big] & \leq  	 \rho_s 	\mathbb{E} \big[\| \nabla f(\tilde{\mathbf{x}}_{s-1}) \|^2 \big] + 2\mu_s \mathbb{E}\big[ f(\tilde{\mathbf{x}}_{s-1}) - f(\tilde{\mathbf{x}}_s) \big] \\
		& \leq \Big[ \prod_{\tau=1}^s \rho_\tau \Big] \| \nabla f(\mathbf{x}_0) \|^2 + 2\sum_{\tau=1}^s  \mu_\tau \mathbb{E}\big[ f(\tilde{\mathbf{x}}_{\tau-1}) - f(\tilde{\mathbf{x}}_\tau) \big] \Big[ \prod_{j=\tau+1}^s \rho_j \Big].
	\end{align*}
	The choice of $\rho_s$ and $\mu_s$ satisfies $\mu_s = \mu_{s-1} \rho_s$, hence the second term is a telescope sum. Thus, we arrive at
	\begin{align*}
		\mathbb{E} \big[ \|  \nabla f(\tilde{\mathbf{x}}_s) \|^2 \big] 
		& \leq \Big[ \prod_{\tau=1}^s \rho_\tau \Big] \| \nabla f(\mathbf{x}_0) \|^2 + 2\mu_0 \Big[ \prod_{\tau=1}^s \rho_\tau \Big] \mathbb{E} \Big[ f(\mathbf{x}_0) - f(\tilde{\mathbf{x}}_s) \Big] \\
		& \leq \Big[ \prod_{\tau=1}^s \rho_\tau \Big] \| \nabla f(\mathbf{x}_0) \|^2 + 2\mu_0 \Big[ \prod_{\tau=1}^s \rho_\tau \Big] \Big( f(\mathbf{x}_0) - f(\mathbf{x}^*) \Big)
	\end{align*}
	which is exactly \eqref{eq.noplateu}. Next, in order to have $\mathbb{E} \big[ \|  \nabla f(\tilde{\mathbf{x}}_s) \|^2 \big] \leq \epsilon $, it suffices to choose $\mu_0 = {\cal O}(1)$ and $\prod_{s=1}^S \rho_s = {\cal O}(\epsilon) $. With the choices of $\eta_s$, $m_s$ and $K_s$, the $s$-th call of SARAH has complexity ${\cal O}\big((n+ \frac{L+\mu_s}{\mu_s}) \ln \frac{1}{\rho_s} \big) ={\cal O}\big((n+ \frac{L}{\mu_s}) \ln \frac{1}{\rho_s} \big)$. As a result, the overall complexity is 
	\begin{align*}
		{\cal O}\bigg( \sum_{s=1}^S \Big[ \big(n+ \frac{L}{\mu_s} \big) \ln \frac{1}{\rho_s}  \Big] \bigg)  < {\cal O}\bigg( \big(n+ \frac{L}{\mu_S} \big) \sum_{s=1}^S  \Big[ \ln \frac{1}{\rho_s}  \Big] \bigg) = {\cal O}\bigg( \big(n+ \frac{L}{\epsilon} \big) \ln \frac{1}{\epsilon}   \bigg)
	\end{align*}
	which completes the proof of the theorem.
\end{proof}

%%%%%%%%%%%%%%%%%%%%%%%%%%%%%%%%%%%%%%%%%%%%%%%%%%%%%%%%%%%%%%%%%%%%%%%%%%%%%%%%%%%%%%%%%%%%%%%%%%%%%%%%%%%%%%%%%%%%%%%%%
\subsection{Proof of Theorem \ref{thm.rsarah_ncvx}}
\begin{proof}
	By definition of $\tilde{f}_s$, we have  $\nabla \tilde{f}_s(\mathbf{x}) = \nabla f(\mathbf{x}) + (\sigma + \theta)(\mathbf{x} - \tilde{\mathbf{x}}_{s-1})$. Then,
	\begin{align*}
		 \| \nabla \tilde{f}_s(\tilde{\mathbf{x}}_s) \|^2 & = \|  \nabla f(\tilde{\mathbf{x}}_s) \|^2 +  (\sigma+\theta)^2 \| \tilde{\mathbf{x}}_s -\tilde{\mathbf{x}}_{s-1} \|^2 + 2(\sigma+\theta) \big\langle \nabla f(\tilde{\mathbf{x}}_s) , \tilde{\mathbf{x}}_s -\tilde{\mathbf{x}}_{s-1} \big\rangle \nonumber \\
		 &\stackrel{(a)}{\geq} \|  \nabla f(\tilde{\mathbf{x}}_s) \|^2 + (\sigma+\theta)^2 \| \tilde{\mathbf{x}}_s -\tilde{\mathbf{x}}_{s-1} \|^2 + 2(\sigma+\theta) \Big[ f(\tilde{\mathbf{x}}_s) - f(\tilde{\mathbf{x}}_{s-1}) - \frac{\sigma}{2} \| \tilde{\mathbf{x}}_s -\tilde{\mathbf{x}}_{s-1} \|^2 \Big]
	\end{align*}
	where (a) is by $\sigma$-bounded nonconvexity of $f$. Rearranging the terms and taking expectation we arrive at 
	\begin{align*}
		\mathbb{E} \big[ \|  \nabla f(\tilde{\mathbf{x}}_s) \|^2 \big] & \leq  	\mathbb{E} \big[ \| \nabla \tilde{f}_s(\tilde{\mathbf{x}}_s) \|^2 \big] - \theta(\sigma+\theta) \mathbb{E} \big[ \| \tilde{\mathbf{x}}_s -\tilde{\mathbf{x}}_{s-1} \|^2\big] + 2 (\sigma + \theta)	\mathbb{E}\big[ f(\tilde{\mathbf{x}}_{s-1}) - f(\tilde{\mathbf{x}}_s) \big]  \\
		&\stackrel{(b)}{\leq} \rho_s \mathbb{E} \big[\| \nabla \tilde{f}_s(\tilde{\mathbf{x}}_{s-1}) \|^2\big] + 2 (\sigma + \theta)	\mathbb{E}\big[ f(\tilde{\mathbf{x}}_{s-1}) - f(\tilde{\mathbf{x}}_s) \big] \\
		& \stackrel{(c)}{=} \rho_s 	\mathbb{E} \big[\| \nabla f(\tilde{\mathbf{x}}_{s-1}) \|^2 \big] + 2 (\sigma + \theta)	 \mathbb{E}\big[ f(\tilde{\mathbf{x}}_{s-1}) - f(\tilde{\mathbf{x}}_s) \big] \nonumber \\
		& \leq  \rho \mathbb{E} \big[\| \nabla f(\tilde{\mathbf{x}}_{s-1}) \|^2 \big] + 2 (\sigma + \theta)	 \mathbb{E}\big[ f(\tilde{\mathbf{x}}_{s-1}) - f(\tilde{\mathbf{x}}_s) \big] 
	\end{align*}
	where (b) is by $\mathbb{E} \big[ \| \nabla \tilde{f}_s(\tilde{\mathbf{x}}_s) \|^2 \big] \leq \rho_s \mathbb{E} \big[\| \nabla \tilde{f}_s(\tilde{\mathbf{x}}_{s-1}) \|^2\big]$; and (c) is by the fact $ \nabla f(\tilde{\mathbf{x}}_{s-1}) = \nabla \tilde{f}_s(\tilde{\mathbf{x}}_{s-1}) $.
	
	Next since $k$ is uniformly chosen from $\{1,\ldots, S\}$, we have
	\begin{align*}
		\mathbb{E} \big[ \|  \nabla f(\tilde{\mathbf{x}}_k) \|^2 \big] & = \frac{1}{S} \sum_{s=1}^S \mathbb{E} \big[ \|  \nabla f(\tilde{\mathbf{x}}_s) \|^2 \big] \nonumber \\
		& \leq  \frac{1}{S}\sum_{s=1}^S \rho \mathbb{E} \big[\| \nabla f(\tilde{\mathbf{x}}_{s-1}) \|^2 \big] +   \frac{1}{S} \sum_{s=1}^S  2 (\sigma + \theta)	 \mathbb{E}\big[ f(\tilde{\mathbf{x}}_{s-1}) - f(\tilde{\mathbf{x}}_s) \big] \nonumber \\
		& \leq \frac{\rho }{S} \| \nabla f(\tilde{\mathbf{x}}_{0}) \|^2 + \frac{1}{S} \sum_{s=1}^{S-1} \rho \mathbb{E} \big[\| \nabla f(\tilde{\mathbf{x}}_s) \|^2 \big] +  \frac{2 (\sigma + \theta)	 \big[ f(\tilde{\mathbf{x}}_0) - f(\mathbf{x}^*) \big]}{S}  \nonumber \\
		& \leq \frac{\rho }{S} \| \nabla f(\tilde{\mathbf{x}}_{0}) \|^2 + \frac{1}{S} \sum_{s=1}^{S} \rho \mathbb{E} \big[\| \nabla f(\tilde{\mathbf{x}}_s) \|^2 \big] +  \frac{2 (\sigma + \theta)	 \big[ f(\tilde{\mathbf{x}}_0) - f(\mathbf{x}^*) \big]}{S}   
	\end{align*}
	Rearranging the terms, we have
	\begin{align*}
		(1-\rho)\mathbb{E} \big[ \|  \nabla f(\tilde{\mathbf{x}}_k) \|^2 \big] \leq \frac{\rho }{S} \| \nabla f(\tilde{\mathbf{x}}_{0}) \|^2 + \frac{2 (\sigma + \theta) \big[ f(\tilde{\mathbf{x}}_0) - f(\mathbf{x}^*) \big]}{S}  
	\end{align*}
	which is equivalent to have 
	\begin{align*}
		\mathbb{E} \big[ \|  \nabla f(\tilde{\mathbf{x}}_k) \|^2 \big] \leq \frac{\rho }{(1-\rho) S} \| \nabla f(\tilde{\mathbf{x}}_{0}) \|^2 + \frac{2 (\sigma + \theta)	  \big[ f(\tilde{\mathbf{x}}_0) - f(\mathbf{x}^*) \big]}{(1-\rho)S}.
	\end{align*}
	To have $\mathbb{E} \big[ \|  \nabla f(\tilde{\mathbf{x}}_k) \|^2 \big] \leq \epsilon$, one can choose $S = {\cal O}\big(\frac{1}{(1-\rho)\epsilon}\big)$. This further leads to a complexity 
	\begin{align*}
		{\cal O} \bigg(  \sum_{s=1}^S \Big(n + \frac{L+\sigma+\theta}{\theta} \Big) K_s \bigg)	 \leq {\cal O}  \bigg(  \frac{1}{\epsilon(1-\rho)}\Big(n + \frac{L+\sigma+\theta}{\theta} \Big) \ln \frac{1}{\rho} \bigg).
	\end{align*}

\end{proof}

%%%%%%%%%%%%%%%%%%%%%%%%%%%%%%%%%%%%%%%%%%%%%%%%%%%%%%%%%%%%%%%%%%%%%%%%%%%%%%%%%%%%%%%%%%%%%%%%%%%%%%%%%%%%%%%%%%%%%%%%%
\subsection{Proof for Corollary \ref{coro.bbsarah_cvx}}
Using similar argument as the proofs in Theorem \ref{thm.single_reg}, we can obtain \eqref{eq.c3} again. This implies that upon choosing $\mu = {\cal O}\big(\frac{\sqrt{\epsilon}}{\|\mathbf{x}_0 - \mathbf{x}^* \|}\big)$, having $\| \nabla \tilde{f}(\mathbf{x}_a) \|^2 \leq \epsilon $ directly implies $\| \nabla f(\mathbf{x}_a) \|^2 	= {\cal O}(\epsilon)$.

Next, using SARAH with BB step size to minimize $\tilde{f}$ will cost ${\cal O}\big( (n+\tilde{\kappa}^2) \ln \frac{\| \nabla f(\mathbf{x}^0) \|^2}{\epsilon}\big)$ to obtain $\mathbb{E}[\| \nabla \tilde{f} (\mathbf{x}) \|^2]\leq \epsilon$ \citep{li2019bb}. Plugging $\tilde{\kappa}$ in, we can obtain the complexity.

The manner for proving other corollaries are exactly the same as Corollary \ref{coro.bbsarah_cvx}, and hence omitted here.

%%%%%%%%%%%%%%%%%%%%%%%%%%%%%%%%%%%%%%%%%%%%%%%%%%%%%%%%%%%%%%%%%%%%%%%%%%%%%%%%%%%%%%%%%%%%%%%%%%%%%%%%%%%%%%%%%%%%%%%%%
\section{Comparison with \citep{ma2018} }\label{apdx.ma2018}
As we shall see in either convex and nonconvex problems, the variant of BB step size in \citep{ma2018} has two parameters require tuning (i.e., $\lambda$ and $m$.) Below we will emphasis the convergence rate of such step size.

\textbf{Convex case:} Another variant of BB step size was studied in \citep{ma2018}, where the following step size is adopted
\begin{equation}\label{eq.bb_stepsize3}
	\eta^{(s)} = \frac{1}{m} \frac{\| \tilde{\mathbf{x}}^{s-1} - \tilde{\mathbf{x}}^{s-2} \|^2}{ \big\langle \tilde{\mathbf{x}}^{s-1} - \tilde{\mathbf{x}}^{s-2}, \nabla f(\tilde{\mathbf{x}}^{s-1}) -  \nabla f(\tilde{\mathbf{x}}^{s-2}) \big\rangle + \lambda \| \tilde{\mathbf{x}}^{s-1} - \tilde{\mathbf{x}}^{s-2} \|^2}. 
\end{equation}
It can be seen that we have $\eta_{\min} = \frac{1}{m (L+\lambda)}\leq \eta^{(s)} \leq \frac{1}{m\lambda} = \eta_{\max}$. As directly analyzing SARAH require extra assumptions as in \citep{nguyen2017,nguyen2018inexact}, we will focus on L2S \citep{li2019l2s} equipping with \eqref{eq.bb_stepsize3}.

Applying similar analysis as \citep[Theorem 1]{li2019l2s}, long as $\eta_{\min} - \eta_{\max}^2 = {\cal O}(\eta_{\min})$ (hence, $\lambda = {\cal O}(1/\sqrt{m})$) and $\eta_{\max} \leq 1/L$, one can get 
\begin{align*}
	 \frac{1}{T}\sum_{t=1}^T \mathbb{E}  \Big[ \|\nabla f(\mathbf{x}_t)\|^2 \Big] \leq {\cal O} \Big( \frac{ f(\mathbf{x}_0) - f(\mathbf{x}^* )}{\eta_{\min} T} +  \frac{m \eta_{\max}^2 }{\eta_{\min}T} \|\nabla f(\mathbf{x}_0) \|^2 \Big)
\end{align*}	
where ${\cal O}$ hides $L$ and constants. Plugging $\eta_{\min}$ and $\eta_{\max}$ in, we arrive at
\begin{align*}
	 \frac{1}{T}\sum_{t=1}^T \mathbb{E}  \Big[ \|\nabla f(\mathbf{x}_t)\|^2 \Big] \leq {\cal O} \Big( \frac{ m(L+\lambda)}{ T} +  \frac{ (L+\lambda)}{ \lambda^2 T}  \Big).
\end{align*}	
Choosing $\lambda = {\cal O}(1/\sqrt{m})$, the convergence rate is ${\cal O}(\frac{m}{T})$. Hence let $ \frac{1}{T}\sum_{t=1}^T \mathbb{E}  \big[ \|\nabla f(\mathbf{x}_t)\|^2 \big] \leq \epsilon$ requires $T = {\cal O}\big( \frac{m}{\epsilon} \big)$, which leads to a complexity of ${\cal O}(n+\frac{n}{\epsilon})$.

\textbf{Nonconvex case:} In \citep{ma2018}, the step size is suggested to be 
\begin{equation}\label{eq.bb_stepsize2}
	\eta^{(s)} = \frac{1}{m} \frac{\| \tilde{\mathbf{x}}^{s-1} - \tilde{\mathbf{x}}^{s-2} \|^2}{ \big| \big\langle \tilde{\mathbf{x}}^{s-1} - \tilde{\mathbf{x}}^{s-2}, \nabla f(\tilde{\mathbf{x}}^{s-1}) -  \nabla f(\tilde{\mathbf{x}}^{s-2}) \big\rangle \big| + \lambda \| \tilde{\mathbf{x}}^{s-1} - \tilde{\mathbf{x}}^{s-2} \|^2}. 
\end{equation}
And it can be seen that $\eta_{\min} = \frac{1}{m (L+\lambda)}\leq \eta^{(s)} \leq \frac{1}{m\lambda} = \eta_{\max}$. Applying similar analysis of \citep{nguyen2019}, it can be seen when $\eta_{\max} = {\cal O}(1/(L\sqrt{m}))$, we have 
\begin{align*}
	 \frac{1}{T}\sum_{t=1}^T \mathbb{E}  \Big[ \|\nabla F(\mathbf{x}_t)\|^2 \Big] = {\cal O} \Big( \frac{1}{\eta_{\min}T} \Big) = {\cal O} \Big( \frac{m}{T} \Big).
\end{align*}
Choosing $m= {\cal O}(n)$ as in \cite{nguyen2019} leads to a complexity of ${\cal O}(\frac{n}{\epsilon})$.

\end{document}